\documentclass[a4paper, 11pt]{article}
\pdfoutput=1
\usepackage{etoolbox,verbatim}

\newtoggle{arxiv}
\toggletrue{arxiv}

\newtoggle{customthms}
\toggletrue{customthms}

\newtoggle{colt}
\togglefalse{colt}

\usepackage{xcolor}
\usepackage{tcolorbox}
\tcbuselibrary{skins,breakable}

\tcbset{
  aibox/.style={
      width=\linewidth,
      top=6pt,
      bottom=0pt,
      left=1.5pt,
      right=1.5pt,
      colframe=black,
      colbacktitle=black,
      enhanced,
      center,
      attach boxed title to top left={yshift=-0.1in,xshift=0.15in},
      boxed title style={boxrule=0pt,colframe=white,},
    }
}
\newtcolorbox{AIbox}[2][]{aibox,title=#2,#1}

\definecolor{primalcolor}{HTML}{960000}
\definecolor{contrarycolor}{HTML}{009696}
\definecolor{primaldarkcolor}{HTML}{660000}
\definecolor{contrarydarkcolor}{HTML}{006666}
\definecolor{irongray}{HTML}{6D6E71}
\definecolor{steelgray}{HTML}{F1F4F7}
\newcommand{\linkcolor}{contrarydarkcolor}
\newcommand{\urlcolor}{contrarydarkcolor}
\newcommand{\citecolor}{contrarydarkcolor}

\newcommand{\thmcolordark}{primaldarkcolor}

\usepackage[T1]{fontenc}

\usepackage{natbib}
\usepackage{amssymb,upgreek,bm}
\usepackage{mathtools}
\usepackage[english]{babel}
\usepackage{multirow,tcolorbox,tikz-cd,turnstile,xspace,xparse}
\usepackage{relsize,breakcites,appendix}
\usepackage{longtable,tablefootnote,booktabs,float,makecell}
\usepackage[normalem]{ulem}
\usepackage{tabu}
\usepackage[shortlabels]{enumitem}
\usepackage{footnote}
\usepackage{boxedminipage}
\usepackage{multirow,nicefrac}
\usepackage{verbatim}
\usepackage{nicematrix}
\usepackage{subcaption}

\usepackage[colorlinks=true,linkcolor=\linkcolor,urlcolor=\urlcolor,citecolor=\citecolor,breaklinks]{hyperref}

\usepackage{prettyref}

\usepackage[capitalise,nameinlink]{cleveref}

\iftoggle{colt}{
    \DeclareRobustCommand{\qed}{
        \usepackage{thmtools}
        \ifmmode \mathqed
        \else
            \leavevmode\unskip\penalty9999 \hbox{}\nobreak\hfill
            \quad\hbox{\qedsymbol}%
        \fi
    }
}
{
    \usepackage{amsthm}

}
\usepackage{thm-restate}

\iftoggle{arxiv}
{
    \usepackage{mathrsfs}
    \usepackage[
        a4paper,
        top=1in,
        bottom=1in,
        left=1in,
        right=1in]{geometry}
    \usepackage[bitstream-charter]{mathdesign}
    
    \usepackage[scaled=0.92]{PTSans}

    \usepackage{fullpage}
    \usepackage[noend]{algpseudocode}
    \usepackage{algorithm}

}
{
    \usepackage{mathrsfs}
}

\DeclareMathAlphabet{\mathbfsf}{\encodingdefault}{\sfdefault}{bx}{n}

\numberwithin{equation}{section}

\iftoggle{arxiv}
{
    \newcommand{\colorbold}[1]{
        \textbf{\textcolor{\thmcolor}{#1}}}}
{
    \newcommand{\colorbold}[1]{\textbf{#1}
    }
}
\iftoggle{arxiv}
{
    
}
{
    
}

\usepackage{thmtools}

\Crefname{equation}{Eq.}{Eqs.}
\Crefname{assumption}{Assumption}{Assumptions}
\Crefname{condition}{Condition}{Conditions}
\Crefname{claim}{Claim}{Claims}
\Crefname{property}{Property}{Properties}
\Crefname{construction}{Construction}{Constructions}

\declaretheoremstyle[
    headformat=\normalfont\textcolor{\thmcolordark}{\bfseries\NAME\,\NUMBER}\NOTE,%
    notefont={\normalfont\textcolor{\thmcolordark}{\bfseries}}, 
    notebraces={}{},
    bodyfont=\normalfont\itshape,
    spaceabove = 6pt,
    spacebelow = 6pt,
    ]{coloredthmversion}

\declaretheoremstyle[
    headformat=\normalfont\textcolor{\thmcolordark}{\bfseries\NAME\,\NUMBER}\NOTE,%
    bodyfont=\normalfont\itshape,
    spaceabove = 6pt,
    spacebelow = 6pt,
    ]{coloredthm}

\declaretheoremstyle[
    headformat=\normalfont\textcolor{\thmcolordark}{\bfseries\NAME\,\NUMBER}\NOTE,%
    bodyfont=\normalfont,
    spaceabove = 6pt,
    spacebelow = 6pt,
    ]{coloreddef}

\iftoggle{customthms}
{
    \theoremstyle{coloredthmversion}
}
{}

\iftoggle{customthms}{
  \theoremstyle{coloredthm}
  \newtheorem{theorem}{Theorem}
  \newtheorem{lemma}{Lemma}[section]
  \newtheorem{corollary}{Corollary}[section]
  \newtheorem{proposition}{Proposition}[section]
}
{}

\newtheorem*{thminformal*}{Informal Theorem}

\iftoggle{customthms}{
    \theoremstyle{coloreddef}
    \newtheorem{definition}{Definition}[section]

    \newtheorem{property}{Property}[section]
}
{
  
}

\newtheorem{assumption}{Assumption}[section]
\newtheorem{condition}{Condition}[section]

\makeatletter
\newcommand{\neutralize}[1]{\expandafter\let\csname c@#1\endcsname\count@}
\makeatother

\iftoggle{customthms}{
    \newtheoremstyle{named}{}{}{\itshape}{}{\bfseries}{}{.5em}{\Cref{#3} {\normalfont (informal)} }{}
    \theoremstyle{named}
    
    \theoremstyle{plain}
}
{}

\newtheorem*{theorem*}{Theorem}
\newtheorem*{lemma*}{Lemma}
\newtheorem*{corollary*}{Corollary}
\newtheorem*{proposition*}{Proposition}
\newtheorem*{claim*}{Claim}
\newtheorem*{fact*}{Fact}
\newtheorem*{observation*}{Observation}
\newtheorem*{definition*}{Definition}
\newtheorem*{remark*}{Remark}
\newtheorem*{example*}{Example}

\def\ddefloop#1{\ifx\ddefloop#1\else\ddef{#1}\expandafter\ddefloop\fi}
\def\ddef#1{\expandafter\def\csname bb#1\endcsname{\ensuremath{\mathbb{#1}}}}
\ddefloop ABCDEFGHIJKLMNOPQRSTUVWXYZ\ddefloop

\def\ddefloop#1{\ifx\ddefloop#1\else\ddef{#1}\expandafter\ddefloop\fi}
\def\ddef#1{\expandafter\def\csname frak#1\endcsname{\ensuremath{\mathfrak{#1}}}}
\ddefloop ABCDEFGHIJKLMNOPQRSTUVWXYZ\ddefloop

\def\ddefloop#1{\ifx\ddefloop#1\else\ddef{#1}\expandafter\ddefloop\fi}
\def\ddef#1{\expandafter\def\csname fr#1\endcsname{\ensuremath{\mathfrak{#1}}}}
\ddefloop ABCDEFGHIJKLMNOPQRSTUVWXYZ\ddefloop

\def\ddefloop#1{\ifx\ddefloop#1\else\ddef{#1}\expandafter\ddefloop\fi}
\def\ddef#1{\expandafter\def\csname eul#1\endcsname{\ensuremath{\EuScript{#1}}}}
\ddefloop ABCDEFGHIJKLMNOPQRSTUVWXYZ\ddefloop

\def\ddefloop#1{\ifx\ddefloop#1\else\ddef{#1}\expandafter\ddefloop\fi}
\def\ddef#1{\expandafter\def\csname scr#1\endcsname{\ensuremath{\mathscr{#1}}}}
\ddefloop ABCDEFGHIJKLMNOPQRSTUVWXYZ\ddefloop

\def\ddefloop#1{\ifx\ddefloop#1\else\ddef{#1}\expandafter\ddefloop\fi}
\def\ddef#1{\expandafter\def\csname b#1\endcsname{\ensuremath{\mathbf{#1}}}}
\ddefloop ABCDEFGHIJKLMNOPQRSTUVWXYZ\ddefloop

\def\ddefloop#1{\ifx\ddefloop#1\else\ddef{#1}\expandafter\ddefloop\fi}
\def\ddef#1{\expandafter\def\csname bhat#1\endcsname{\ensuremath{\hat{\mathbf{#1}}}}}
\ddefloop ABCDEFGHIJKLMNOPQRSTUVWXYZ\ddefloop

\def\ddefloop#1{\ifx\ddefloop#1\else\ddef{#1}\expandafter\ddefloop\fi}
\def\ddef#1{\expandafter\def\csname btil#1\endcsname{\ensuremath{\tilde{\mathbf{#1}}}}}
\ddefloop ABCDEFGHIJKLMNOPQRSTUVWXYZ\ddefloop

\def\ddefloop#1{\ifx\ddefloop#1\else\ddef{#1}\expandafter\ddefloop\fi}
\def\ddef#1{\expandafter\def\csname bst#1\endcsname{\ensuremath{\mathbf{#1}^\star}}}
\ddefloop ABCDEFGHIJKLMNOPQRSTUVWXYZ\ddefloop

\def\ddefloop#1{\ifx\ddefloop#1\else\ddef{#1}\expandafter\ddefloop\fi}
\def\ddef#1{\expandafter\def\csname bst#1\endcsname{\ensuremath{\mathbf{#1}^\star}}}
\ddefloop abcdeghijklmnopqrstuvwxyz\ddefloop

\def\ddefloop#1{\ifx\ddefloop#1\else\ddef{#1}\expandafter\ddefloop\fi}
\def\ddef#1{\expandafter\def\csname bhat#1\endcsname{\ensuremath{\hat{\mathbf{#1}}}}}
\ddefloop abcdefghijklmnopqrstuvwxyz\ddefloop

\def\ddefloop#1{\ifx\ddefloop#1\else\ddef{#1}\expandafter\ddefloop\fi}
\def\ddef#1{\expandafter\def\csname b#1\endcsname{\ensuremath{\mathbf{#1}}}}
\ddefloop abcdeghijklnopqrstuvwxyz\ddefloop

\def\ddefloop#1{\ifx\ddefloop#1\else\ddef{#1}\expandafter\ddefloop\fi}
\def\ddef#1{\expandafter\def\csname barb#1\endcsname{\ensuremath{\bar{\mathbf{#1}}}}}
\ddefloop abcdefghijklmnopqrstuvwxyz\ddefloop

\def\ddef#1{\expandafter\def\csname c#1\endcsname{\ensuremath{\mathcal{#1}}}}
\ddefloop ABCDEFGHIJKLMNOPQRSTUVWXYZ\ddefloop
\def\ddef#1{\expandafter\def\csname h#1\endcsname{\ensuremath{\widehat{#1}}}}
\ddefloop ABCDEFGHIJKLMNOPQRSTUVWXYZ\ddefloop
\def\ddef#1{\expandafter\def\csname hc#1\endcsname{\ensuremath{\widehat{\mathcal{#1}}}}}
\ddefloop ABCDEFGHIJKLMNOPQRSTUVWXYZ\ddefloop
\def\ddef#1{\expandafter\def\csname t#1\endcsname{\ensuremath{\widetilde{#1}}}}
\ddefloop ABCDEFGHIJKLMNOPQRSTUVWXYZ\ddefloop
\def\ddef#1{\expandafter\def\csname tc#1\endcsname{\ensuremath{\widetilde{\mathcal{#1}}}}}
\ddefloop ABCDEFGHIJKLMNOPQRSTUVWXYZ\ddefloop

\DeclareMathOperator*{\argmax}{arg\,max}

\newcommand{\ballkr}[1][r]{\cB_{k}(r)}

\DeclareMathSymbol{\shortminus}{\mathbin}{AMSa}{"39}

\makeatletter
\newcommand\addtometadatalist[5][]{%
    \begingroup
    \if\relax#3\relax\def\sep{}\else\def\sep{#5}\fi
    \let\protect\@unexpandable@protect
    \xdef#3{\expandafter{#3}\sep #4[#1]{#2}}%
    \endgroup
}

\newcommand\metadatalist{}
\newcommand\metadataformat[2][]{{\small \textbf{#1:} #2}}
\newcommand\metadata[2][]{\addtometadatalist[#1]{#2}{\metadatalist}{\metadataformat}{\\}}
\makeatother

\newcommand{\paperwebsite}[1]{\metadata[Website]{\url{#1}}}
\newcommand{\papercode}[1]{\metadata[Code]{\url{#1}}}
\newcommand{\paperdocs}[1]{\metadata[Documentation]{\url{#1}}}
\newcommand{\paperblog}[1]{\metadata[Blog]{\url{#1}}}
\renewcommand\date[1]{\metadata[Date]{#1}}

\usepackage{graphicx}
\usepackage{adjustbox}
\usepackage{wrapfig}

\usepackage[skip=0.5em plus 0.2em minus 0.1em]{parskip}
\makeatletter
\patchcmd{\quotation}{\listparindent 1.5em}{\listparindent\z@}{}{\PackageWarning{specific_macros}{quotation indent patch failed}}
\patchcmd{\quotation}{\parsep \z@ \@plus\p@}{\parsep\parskip}{}{\PackageWarning{specific_macros}{quotation parsep patch failed}}
\makeatother

\hypersetup{
   pdftitle={SWAP: Test-Time Steerable Humanoid Whole-Body Collision Avoidance},
   pdfauthor={Zeji Yi, Guanqi He, Xiaofeng Guo, Guannan Qu, Pieter Abbeel, Yan Duan, Karen Liu, Guanya Shi},
   pdfsubject={Humanoid Robotics, Diffusion Models, Obstacle Avoidance},
   pdfkeywords={Humanoid, Diffusion, Whole-Body, Locomotion, Obstacle Avoidance}
}

\newcommand{\method}{SWAP}

\usepackage{preamble/color-edits}
\addauthor{guanya}{blue}
\addauthor{zeji}{orange}
\addauthor{guanqi}{purple}

\newcommand{\ignore}[1]{}

\iftoggle{arxiv}
{
\input{preamble/arxiv_title}
}
{}

\title{SWAP: Test-Time Steerable Humanoid Whole-Body Collision Avoidance}

\author{
Zeji Yi$^{1,2*}$, Guanqi He$^{2}$, Xiaofeng Guo$^{2}$, Guannan Qu$^{1}$, \\
Pieter Abbeel$^{2}$, Yan Duan$^{2}$, Karen Liu$^{2}$, Guanya Shi$^{1,2}$ \\
\vspace{-.3em}
\rule{.38\textwidth}{.7pt}
\\
\footnotesize
$^{1}$Carnegie Mellon University \quad $^{2}$Amazon FAR \\
\footnotesize
$^{*}$Work done at Amazon FAR
}

\date{\today}
\paperwebsite{https://iscoyizj.github.io/SWAP-steerable-humanoid-oa-website/}

\begin{document}

\begin{tcolorbox}[
    colback=steelgray, colframe=irongray,
    boxrule=0pt,
    before upper={\setlength{\parskip}{0pt}},
    arc=2mm
  ]
  \maketitle
  \tcbline
  \begin{minipage}[t]{1.0\linewidth}
    \centering
    \captionsetup{font=footnotesize}
    \includegraphics[width=\linewidth]{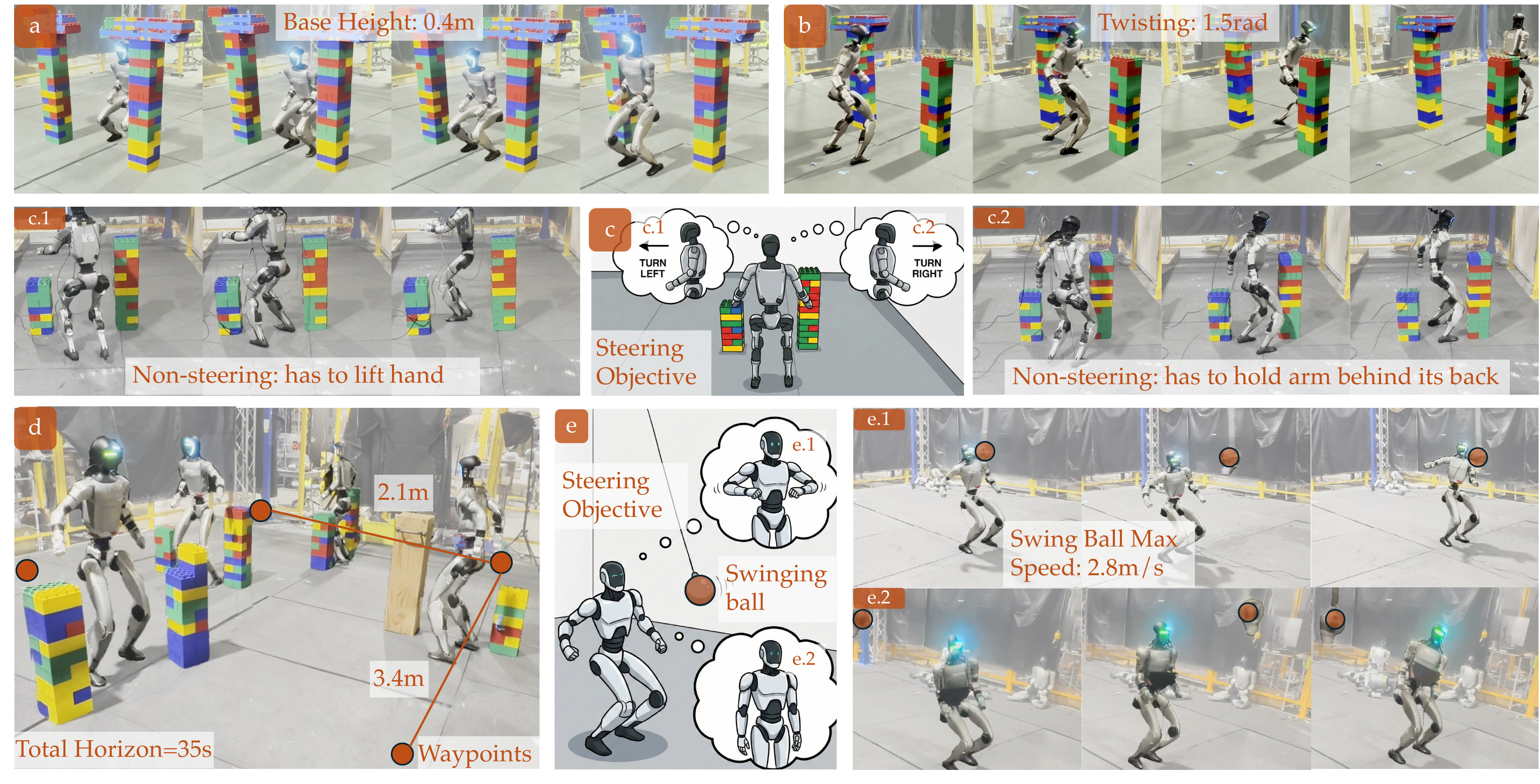}
    \captionof{figure}{%
        \textbf{\method{} enables whole-body obstacle avoidance for humanoid robots.}
        Unlike prior methods that adjust only footsteps, our hierarchical diffusion framework generates coordinated full-body maneuvers including ducking, sidestepping, and twisting.
        \textbf{(a)} Ducking gait with base height as low as 0.40\,m to pass under low-clearance obstacles.
        \textbf{(b)} Torso twisting maneuver with waist rotation up to 1.5\,rad to navigate through narrow gaps.
        \textbf{(c.1--c.2)} Sidestepping behavior with different steering objectives: (c.1) faces left while (c.2) faces right to avoid the obstacles. Notably, the arm lifting and retraction emerges naturally from the diffusion policy rather than explicit steering, demonstrating that collision avoidance generalizes across different objectives.
        \textbf{(d)} Long-horizon traversal through a cluttered environment with four obstacle sets, demonstrating reliable and consistent avoidance behavior.
        \textbf{(e.1--e.2)} Dynamic obstacle avoidance against a swinging ball on a rope: (e.1) steers to raise arms while (e.2) steers to lower arms to evade the moving obstacle.
    }
    \label{fig:teaser}
\end{minipage}

  \tcbline
  \makeatletter
  \ifdefempty{\metadatalist}{}{\metadatalist\par}
  \makeatother
\end{tcolorbox}

\begin{abstract}

Humanoid robots operating in human-centric environments must coordinate the entire body to traverse tight, cluttered spaces while maintaining balance. Existing methods often struggle to jointly consider dynamic stability and collision avoidance from high-dimensional perceptual inputs. Moreover, for humanoids, safe behavior is inherently multimodal: the same obstacle may admit multiple valid traversals, which is difficult to represent with a single deterministic policy. To address these challenges, we propose \method{} (\textit{Steerable Whole-body Avoidance Policy}), a hierarchical framework that separates collision-aware decision making from stabilization. A high-level, perception-conditioned diffusion policy generates multimodal, collision-aware motion proposals, while a low-level whole-body RL controller tracks these proposals and maintains dynamic balance. To efficiently train the high-level policy, we obtain scalable supervision via hindsight labeling: we roll out the low-level policy under randomized commands, place obstacles only in the space that each executed motion did not sweep, so the motion is collision-free by construction, and re-render the perception the robot would have seen. At deployment, Monte Carlo sampling with simple cost signals steers the diffusion model toward safe, task-aligned behaviors without heavy reward tuning.

Experiments in simulation and on a real Unitree G1 humanoid show reliable avoidance with static and dynamic obstacles. Test-time steering further enables customization, improving targeted rewards (e.g., speed, orientation, or posture) by \textbf{34\%} on average while maintaining a high collision avoidance rate. Together, these results support a practical solution to humanoid whole-body collision avoidance with test-time steerability.

Project website:\newline \url{https://iscoyizj.github.io/SWAP-steerable-humanoid-oa-website/}

\end{abstract}

\section{Introduction}
\label{sec:introduction}

Humanoid robots possess the kinematic flexibility to navigate cluttered, human-centric environments in ways that simpler platforms cannot: ducking under obstacles, twisting through narrow gaps, and sidestepping around barriers while maintaining balance. Exploiting this flexibility requires \textit{whole-body coordination}---orchestrating not just footsteps, but whole-body posture in concert~\citep{radosavovic2024realworld, cheng2024expressive}. This capability represents both the defining advantage of the humanoid form factor and its central control challenge.

Yet, achieving reliable whole-body obstacle avoidance remains fundamentally difficult precisely because of the high degree-of-freedom, inherently unstable nature of humanoid robots. A successful policy must simultaneously satisfy dynamic feasibility, control stability, and collision avoidance---a combination that existing methods struggle to achieve.
\textit{Reinforcement learning} struggles with the multimodality of avoidance: many valid traversals exist, but reward-driven optimization collapses to a narrow mode. Shaped, multi-term objectives introduce spurious local minima and reward-sensitive behavior as shown in both humanoid and other legged robots~\citep{wang2025safecomfortable, he2024abs}.
\textit{Model-based planners} face analogous issues---whole-body planning with hand-crafted costs is nonconvex and initialization-sensitive, often yielding stiff, conservative motions that underutilize the humanoid's flexibility~\citep{spark2025, chomp2013}.
\textit{Behavior cloning} can learn natural motions from demonstrations but requires large-scale data; teleoperation on humanoids is difficult and does not scale. In manipulation, imitation-based generative policies can represent multimodal manipulation while achieving collision avoidance, but still depend on demonstration coverage~\citep{li2024lano3dp, rail2024, ze20243ddp, chi2023diffusionpolicy}.
Moreover, policies trained under any of these paradigms cannot easily adapt to changing objectives  
at deployment---any shift in user preference (e.g., prioritizing certain gestures) typically demands retraining.   

In this work, we propose \method{} (\textit{Steerable Whole-body Avoidance Policy}), a hierarchical framework that decouples high-level collision-aware decision making from low-level locomotion control. This separation is key to resolving the stability and collision avoidance dilemma that plagues joint optimization: the low-level RL controller focuses solely on stable trajectory tracking, while the high-level planner handles collision-aware motion generation. For the high-level policy, inspired by diffusion-based whole-body motion synthesis~\citep{huang2025diffusecloc, liao2025beyondmimic},
we adopt a Diffusion Transformer (DiT)~\citep{peebles2023dit} conditioned on LiDAR observations and trained on trajectories derived from the LAFAN1 human motion dataset~\citep{harvey2020lafan1}. This motion prior helps the planner preserve diverse, human-like whole-body behavior. 

To obtain collision labels without teleoperation, we use a hindsight labeling \citep{andrychowicz2017her, gu2024rt_trajectory} pipeline in simulation. We roll out the low-level controller under randomized commands, then run geometric collision checking along the executed motion by testing the robot's swept volume against the environment, producing collision-free labels for each trajectory. This yields scalable collision-aware supervision while avoiding reward shaping: collision enters only as a label on motion-prior trajectories, which helps retain multimodality and naturalness while learning to avoid obstacles. This self-supervised pipeline provides a more scalable way of collecting data and bypasses the need for teleoperation. At deployment, we further steer the diffusion sampling process using Monte Carlo sampling with lightweight cost signals.
Together, these components transform avoidance from a fixed reactive mapping into a flexible, steerable generative task, enabling zero-shot adaptation to new objectives without retraining.

The key capability of \method{} is not any single motion primitive, such as ducking or sidestepping, but the ability to produce collision-aware whole-body proposals that remain \emph{steerable at deployment}. A user can plug in analytical objectives such as speed, heading, height, or posture, or use a learned collision-aware Q-function, and the same trained diffusion prior can adapt by selecting among its sampled trajectory modes. This distinguishes \method{} from approaches that require a new reward, curriculum, or demonstration set whenever the deployment objective changes.

Our main contributions are three-fold:
\begin{enumerate}
    \item A hierarchical framework combining a perception-conditioned DiT with an RL-based controller for whole-body obstacle avoidance, validated on the Unitree G1 in simulation and real-world deployment.

    \item A hindsight labeling pipeline that uses synthetic environments to generate collision-aware supervision for the high-level diffusion policy at scale, without teleoperation.

    \item A lightweight test-time steering mechanism that adapts to new objectives without retraining, improving targeted metrics by \textbf{34\%} on average.
\end{enumerate}

We validate \method{} extensively in simulation and on a Unitree G1 humanoid equipped with LiDAR sensors in the real world. Our experiments demonstrate reliable avoidance of both static and dynamic obstacles, with the robot exhibiting diverse whole-body maneuvers including ducking, sidestepping, and twisting. Together, these results establish a practical framework for humanoid whole-body collision avoidance that integrates naturally with downstream navigation stacks.

\section{Related Work}
\label{sec:related_work}

\begin{figure}[t]
    \centering
    \includegraphics[width=\linewidth]{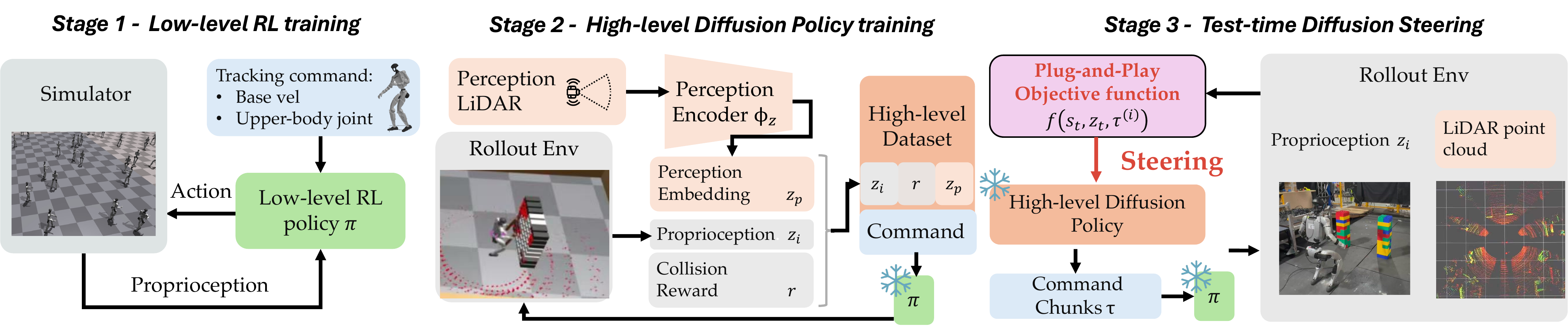}
    \caption{\textbf{Overview of the \method{} framework.} The high-level diffusion policy generates collision-aware whole-body trajectories from perception observations, which are tracked by a low-level RL locomotion controller. }
    \label{fig:system_overview}
\end{figure}

Our approach lies at the intersection of learning-based locomotion, generative models for control, and safe humanoid navigation. We review relevant work in each area.
\subsection{Learning-Based Humanoid Locomotion}
Humanoid locomotion has seen significant progress through reinforcement learning (RL), moving from basic bipedal balance~\citep{radosavovic2024realworld} to perceptive locomotion in rough terrains~\citep{zhuang2024parkour}.
Recent hierarchical approaches, such as HWC-Loco~\citep{lin2025hwcloco}, decompose the problem into high-level planning and low-level control to achieve robust gait. Similarly, HumanPlus~\citep{fu2024humanplus} demonstrates the effectiveness of shadowing and imitation learning for whole-body coordination. 
Generative motion priors, such as NaturalGMP~\citep{zhang2025naturalgmp}, further enhance motion naturalness without hand-crafted style rewards. While these research efforts provide a solid foundation for mobility, the integration of these motor skills into active collision avoidance in complex, cluttered environments remains a central challenge.
\subsection{Diffusion Models for Motion Planning}
Diffusion models have emerged as powerful generative priors for robot policy learning. In manipulation, diffusion policies have shown strong performance~\citep{chi2023diffusionpolicy, ze20243ddp}, including on humanoid robots~\citep{ze2024humanoidmanipulation}. For navigation, NavDP~\citep{cai2025navdp} combines a diffusion policy with privileged information guidance and a safety critic, achieving successful sim-to-real navigation transfer.

For humanoid control specifically, recent works leverage diffusion as a motion prior. 
DreamControl~\citep{kalaria2025dreamcontrol} uses a diffusion prior over human motions to help an RL policy solve long-horizon humanoid tasks. Most relevant to our work, BeyondMimic~\citep{liao2025beyondmimic} demonstrates that guided diffusion can achieve zero-shot task adaptation, including obstacle avoidance, by applying cost-guided sampling to a diffusion controller. While these frameworks excel at generating feasible behaviors, steering them toward task-specific safety in cluttered environments often requires expert data or costly online optimization; this motivates our hindsight collision labeling as an efficient supervision signal for systematic whole-body obstacle avoidance.
\subsection{Safe Navigation for Legged Robots}
Ensuring safety in humanoid navigation remains challenging, as it requires satisfying dynamic feasibility, control stability, and collision avoidance simultaneously. Control barrier functions (CBFs)~\citep{ames2019cbf} provide formal safety guarantees but can be conservative. \citet{wang2025safecomfortable} formulate humanoid locomotion as a constrained MDP with CBF costs, achieving safe and comfortable walking in crowds with the perception module of Omni-Perception~\citep{wang2025omniperception}. However, such constraint-based approaches often require careful tuning and may not generalize across diverse environments. 

To address these limitations, reactive approaches like HumanoidPF~\citep{xue2026humanoidpf} encode humanoid-obstacle relationships into an artificial potential field to guide RL policies. While effective for narrow passage traversal, such methods typically rely on explicit geometric mappings and require intensive reward engineering to balance stability with safety. Other learning-based approaches to safety include training with collision penalties, using safety critics as seen in NavDP~\citep{cai2025navdp}, or employing curriculum learning. For example, \citet{kim2024armor} propose ARMOR, which uses an inverse hindsight strategy by synthetically placing obstacles around known collision-free human motions to guarantee safe demonstrations.

In contrast, our approach introduces a physically grounded hindsight labeling strategy derived from the robot's own swept volume. This enables the policy to internalize diverse, multimodal avoidance maneuvers consistent with its kinematic limits, eliminating the need for manual reward shaping or external expert demonstrations.

\colorbold{Positioning.} Prior systems provide important pieces of the humanoid avoidance stack, but none combines whole-body obstacle avoidance with plug-and-play test-time objectives. Motion-imitation and generative locomotion methods provide natural behavior priors but do not directly solve perception-conditioned collision avoidance. ARMOR~\citep{kim2024armor} and HumanoidPF~\citep{xue2026humanoidpf} target perception-driven avoidance, but their deployment behavior is tied to the trained objective or hand-designed potential field. Model-based planners such as SPARK~\citep{spark2025} can optimize explicit costs, but scaling them to real-time whole-body humanoid motion with learned perception remains difficult. \method{} instead learns a multimodal diffusion prior over feasible whole-body commands and adapts online by scoring sampled trajectories, preserving the hierarchy while avoiding task-specific retraining.

\section{Steerable Hierarchical Policy for Whole-Body Obstacle Avoidance}
\label{sec:steerable_hierarchical_policy}

We consider a humanoid robot navigating cluttered, human-centric environments. The robot must reach goal locations while avoiding collisions using its full body---not only via footstep adjustments, but through coordinated motions of the torso, arms, and legs. Appendix~\ref{appendix:notation} summarizes our notation.

\colorbold{Why Hierarchical Control?} Directly generating joint-level commands with diffusion models poses challenges for humanoid control: the extensive computation of iterative denoising struggles to meet high-frequency requirements (30--50\,Hz), and end-to-end policies must simultaneously learn collision avoidance and dynamic stability---competing objectives that are difficult to balance. We adopt a hierarchical decomposition: a high-level diffusion planner generates trajectory commands at lower frequency, while a low-level RL controller runs at high frequency for stable execution. As we show in Section~\ref{sec:experiments}, this separation yields significantly stronger performance compared to end-to-end diffusion baselines.

\colorbold{Hierarchical Control Formulation.} Let $\mathbf{o}_t^{\text{obs}}$ denote an exteroceptive observation of nearby obstacles (e.g., LiDAR point clouds), and $\mathbf{s}_t$ the proprioceptive state. A perception encoder first processes the observation with temporal context to produce a perception embedding $\mathbf{z}_t = \text{Enc}(\mathbf{o}_t^{\text{obs}}, \mathbf{z}_{t-1})$. The high-level policy $\pi_H$ then outputs a trajectory command over a horizon $H$:
\begin{equation}
    \boldsymbol{\tau}_{t:t+H} = \pi_H(\mathbf{z}_t, \mathbf{s}_t)
\end{equation}
where $\boldsymbol{\tau}_{t:t+H}$ represents commanded waypoints for the robot's base (position, velocity, height) and upper-body DoFs---not full joint-level commands. The low-level controller $\pi_L$ tracks these waypoints and outputs motor commands:
\begin{equation}
    \mathbf{a}_t = \pi_L(\boldsymbol{\tau}_t, \mathbf{q}_t, \dot{\mathbf{q}}_t)
\end{equation}
This separation allows $\pi_H$ to specify desired base and upper-body motion for collision avoidance, while $\pi_L$ determines the full joint trajectories needed to achieve stable tracking. Figure~\ref{fig:system_overview} illustrates this architecture.

\subsection{Low-Level RL Policy Pretraining}
\label{sec:low_level_pretraining}

The low-level controller is trained via reinforcement learning to track high-level commands while maintaining dynamic stability. To decouple navigation from stabilization, the policy is proprioceptive-only, focusing exclusively on locomotion and balance without exteroceptive obstacle information. Once trained, the policy is frozen to serve as a fixed action space for the high-level planner.

\colorbold{Observation and Action Space.} Following the decoupled control paradigm in FALCON~\citep{zhang2025falcon}, we design a hierarchical command space to separate locomotion from upper-body posturing. The policy $\pi_L$ receives an observation $\mathbf{o}_t^{L}$ consisting of the target trajectory waypoint, the robot's proprioceptive state (joint positions, velocities, and orientation), and the previous action. It outputs target joint positions $\mathbf{a}_t$ converted to motor torques via a PD controller. 

To ensure the policy masters a diverse range of whole-body configurations, we employ a hybrid command sampling strategy during training. For the lower body, base linear and angular velocity commands $\mathbf{v}_t^{b*}$ are sampled from a random distribution to promote robust locomotion. For the upper body, reference joint positions $\mathbf{p}_t^{u*}$ are randomly sampled from the LAFAN1 human motion dataset~\citep{harvey2020lafan1}. This exploration is crucial for generating the rich swept volume data used for subsequent hindsight labeling.

\colorbold{Reward Design.} The policy is optimized via a reward $r = w_{\text{track}} \cdot r_{\text{track}} + r_{\text{locomotion}}$ which balances tracking accuracy with fundamental stability. The tracking term is defined as:
\begin{equation}
    r_{\text{track}} = \exp\!\left(
    \begin{aligned}
        &-\tfrac{1}{\sigma_v^2}\|\mathbf{v}_t^{b} - \mathbf{v}_t^{b*}\|_2^2
        -\tfrac{1}{\sigma_p^2}\|\mathbf{p}_t^{b} - \mathbf{p}_t^{b*}\|_2^2 \\
        &-\tfrac{1}{\sigma_h^2}(h_t - h_t^*)^2
        -\tfrac{1}{\sigma_u^2}\|\mathbf{p}_t^{u} - \mathbf{p}_t^{u*}\|_2^2
    \end{aligned}
    \right).
\end{equation}
The term $r_{\text{locomotion}}$ penalizes body tilting, base angular velocities, and joint accelerations to ensure smooth movement. 
\subsection{Hindsight Labeling for High-Level Diffusion Policy}
\label{sec:hindsight_labeling}

Training the high-level diffusion policy requires perception--trajectory pairs. We obtain supervision via hindsight labeling using executed trajectory replay: we first generate diverse motions by rolling out randomized commands through the low-level controller, then retrospectively construct a labeled obstacle representation from the swept volume of the executed motion, and finally re-render perception observations along that trajectory.

\colorbold{Procedure.} We first \textbf{roll out commands from a human motion prior} in simulation: we sample upper-body motion segments from LAFAN1 and concatenate them with randomized base waypoints (position/heading/height targets) to form a desired high-level command sequence $\{\boldsymbol{\tau}_t\}_{t=0}^{T}$. Executing $\{\boldsymbol{\tau}_t\}$ with the frozen low-level controller yields an actual joint trajectory $\{\mathbf{q}_t\}_{t=0}^{T}$. We use this executed trajectory, not the commanded reference, to compute the swept volume---the set of workspace points occupied by the robot body over time. Thus, the high-level dataset already reflects low-level tracking errors and physically feasible motions. Formally, define
\begin{equation}
    \mathcal{S} \;=\; \bigcup_{t=0}^{T} \mathcal{R}(\mathbf{q}_t),
\end{equation}
where $\mathcal{R}(\mathbf{q}_t) \subset \mathbb{R}^3$ is the robot body occupancy at configuration $\mathbf{q}_t$. We \textbf{label obstacles from the un-swept space} by sampling obstacle configurations from the complement $\mathcal{W} \setminus \mathcal{S}$, where $\mathcal{W}$ is the workspace region. Intuitively, $\mathcal{S}$ is the space traversed by the robot, so any obstacle set $\mathcal{O}_{\text{hind}} \subseteq \mathcal{W} \setminus \mathcal{S}$ forms a collision-free training pair for that executed trajectory.

\colorbold{Data Augmentation.} From each base obstacle configuration $\mathcal{O}_{\text{hind}}$, we generate augmented variants for two purposes. (1)~\textit{Augmentation for positive data}: we randomly remove subsets of obstacles from $\mathcal{O}_{\text{hind}}$. The smaller obstacle set still lies in the un-swept space $\mathcal{W} \setminus \mathcal{S}$, so the original trajectory remains collision-free. This increases the diversity of safe training pairs for the diffusion policy. (2)~\textit{Augmentation for negative data}: we add obstacles inside the swept volume $\mathcal{S}$, making the same kinematic rollout unsafe under the new obstacle configuration. These negative samples are used for training the objective function $f$ for test-time steering (\cref{sec:steering}), not for the diffusion policy. Importantly, both augmentation strategies maintain physical plausibility by ensuring all obstacles are grounded (no floating obstacles) and respect spatial coherence. 

\begin{wrapfigure}{r}{0.4\linewidth}
    \centering
    \vspace{-1.2em}
    \includegraphics[width=\linewidth]{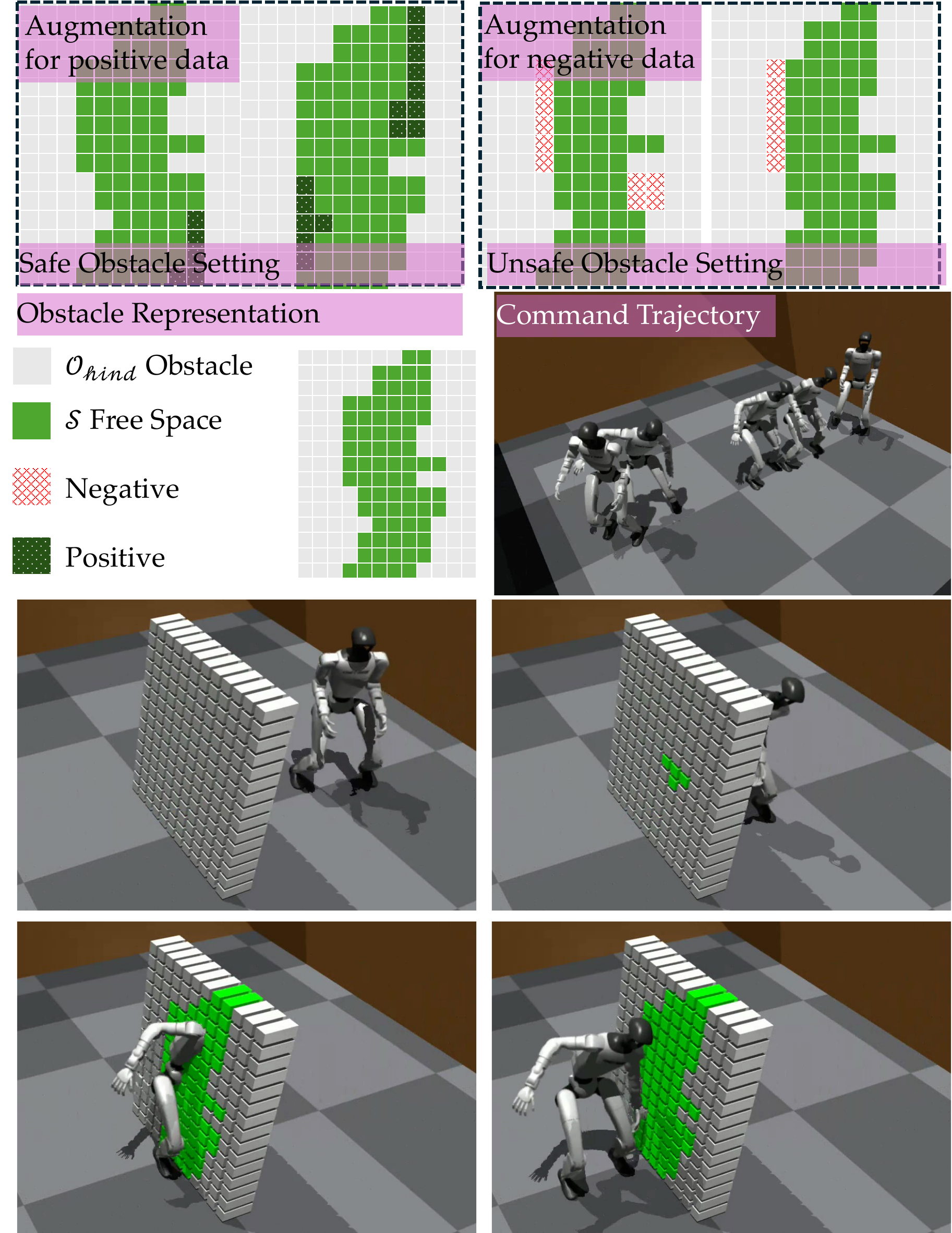}
    \caption{\textbf{Hindsight labeling pipeline with kinematic replay.} Randomized rollouts generate diverse motions; the swept volume of the replayed trajectory is used to label a conservative obstacle representation, and perception is re-rendered along the replay for training the diffusion policy.}
    \label{fig:hindsight}
    \vspace{-1em}
\end{wrapfigure}
Finally, we \textbf{re-render perception} by replaying the kinematic trajectory and synthesizing observations $\mathbf{o}_t^{\text{obs}}$ (via ray casting) against the augmented obstacle configurations. Random Gaussian noise is added to the re-rendered perception to simulate the real-world noise.
This hindsight construction avoids manual annotation: labels and perception are generated automatically from the robot's motion trace, while augmentation provides diverse supervision for both the diffusion policy and the steering objective function $f$. 

\colorbold{Dataset and Filtering.} Our hindsight dataset contains 2,356 executed rollouts. We apply 100$\times$ positive and negative augmentation and remove physically implausible obstacle configurations (e.g., an obstacle floating without support), which leaves 78,900 perception--trajectory pairs. Upper-body commands are sampled from LAFAN1 motion segments, while base position, heading, velocity, and height targets are sampled independently to cover diverse whole-body commands. We filter rollouts in which the low-level controller falls or exhibits large tracking error; this removes about 3\% of the collected data. This filtering biases the high-level planner toward motions that the low-level controller can actually execute, which is desirable for a hierarchical system because infeasible high-level plans would otherwise fail at deployment.

\colorbold{Domain Randomization.} We randomize the wall configuration in Figure~\ref{fig:hindsight} by perturbing its yaw within $[-20^\circ, 20^\circ]$ and shifting its center by $\pm 0.5\,$m. Commands are sampled from the LAFAN1 human motion dataset, and we inject Gaussian noise into both the upper body command stream and the base waypoint targets (e.g., initial position, velocity, and height) to improve robustness.
\subsection{Perception Module}
\label{sec:perception_module}

\begin{figure}[t]
    \centering
    \includegraphics[width=0.9\linewidth]{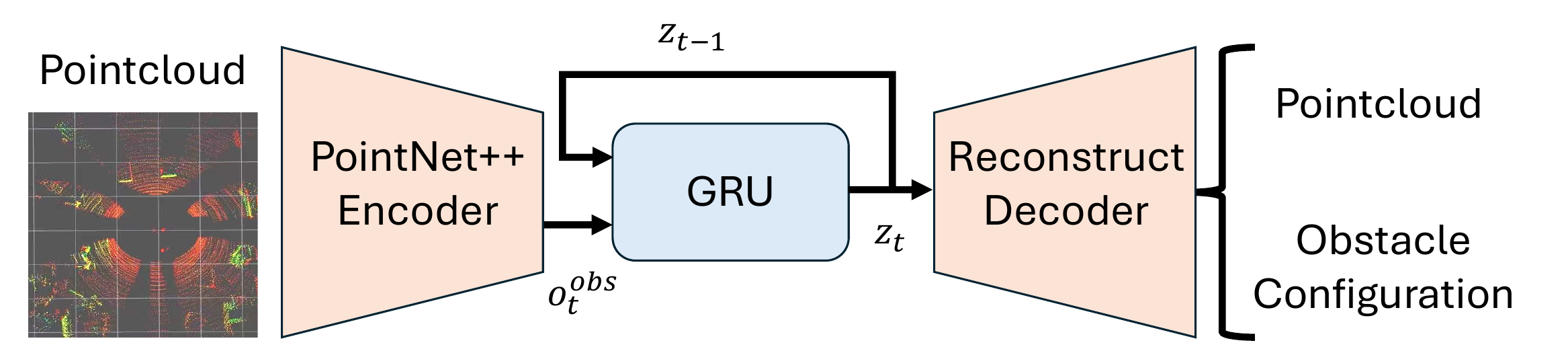}
    \caption{\textbf{Perception encoder architecture}. Point clouds are encoded via PointNet++ and a GRU into a perception embedding $\mathbf{z}_t$. To ensure spatial awareness, $\mathbf{z}_t$ is supervised via reconstruction of point clouds and obstacle configuration.}
    \label{fig:perception_encoder}
\end{figure}

The robot is equipped with (i) an exteroceptive perception sensor providing $\mathbf{o}_t^{\text{obs}}$ (e.g., LiDAR point clouds or depth images), and (ii) proprioception including joint positions $\mathbf{q}_t$, velocities $\dot{\mathbf{q}}_t$, and IMU measurements.

\colorbold{Perception Encoder with Temporal Context.} The point cloud observation $\mathbf{o}_t^{\text{obs}}$ is processed by an encoder network that incorporates temporal context (Figure~\ref{fig:perception_encoder}). The encoder takes as input both the current observation and the previous perception embedding $\mathbf{z}_{t-1}$:
\begin{equation}
    \mathbf{z}_t = \text{Enc}(\mathbf{o}_t^{\text{obs}}, \mathbf{z}_{t-1})
\end{equation}
This recurrent structure enables the encoder to capture temporal patterns in obstacle motion and maintain spatial awareness across frames, with the embedding $\mathbf{z}_t$ serving as both the output for the current step and the temporal context for the next.

The perception embedding $\mathbf{z}_t$ is concatenated with the proprioceptive state $\mathbf{s}_t$ to form the conditioning vector for the high-level diffusion policy, whose input is $\mathbf{c} = (\mathbf{z}_t, \mathbf{s}_t)$ and output is a future action chunk $\boldsymbol{\tau}_{t:t+H}$. In our implementation, we use LiDAR point clouds processed by a PointNet-style encoder~\citep{wang2025omniperception, wang2025safecomfortable}, though the framework is agnostic to the specific perception modality.

\subsection{High-Level Diffusion Policy Training}
\label{sec:diffusion_training}

The high-level policy is a conditional diffusion model that maps perception-conditioned context to whole-body trajectory commands. Unlike approaches that output simple velocity commands, our planner generates rich trajectory information (e.g., body poses / waypoints over a prediction horizon).

\colorbold{Diffusion Model Architecture.} We adopt a denoising diffusion probabilistic model (DDPM). Given a noisy trajectory $\boldsymbol{\tau}^{(k)}$ at diffusion step $k$, here we omit the subscript $t:t+H$ for brevity, the model predicts a denoising direction:
\begin{equation}
    \boldsymbol{\epsilon}_\theta(\boldsymbol{\tau}^{(k)}, k, \mathbf{c}) \rightarrow \boldsymbol{\tau}^{(k-1)}
\end{equation}
In practice, the conditioning uses a 5-step history: $\mathbf{c} = (\mathbf{z}_{t}, \mathbf{s}_{t-4:t})$ consists of the perception embedding $\mathbf{z}_{t}$ from the encoder (Section~\ref{sec:perception_module}) and the proprioceptive state $\mathbf{s}_{t-4:t}$.

\colorbold{Training with Hindsight Labels.} We train the diffusion policy on the hindsight-labeled dataset (Section~\ref{sec:hindsight_labeling}) using the standard DDPM objective. Given a clean trajectory $\boldsymbol{\tau}_0$ from the dataset and Gaussian noise $\boldsymbol{\epsilon} \sim \mathcal{N}(\mathbf{0}, \mathbf{I})$, the noisy trajectory at diffusion step $k$ is constructed as:
\begin{equation}
    \boldsymbol{\tau}^{(k)} = \sqrt{\bar{\alpha}_k} \boldsymbol{\tau}_0 + \sqrt{1 - \bar{\alpha}_k} \boldsymbol{\epsilon},
\end{equation}
where $\bar{\alpha}_k = \prod_{i=1}^{k} \alpha_i$ is the cumulative noise schedule. The model is trained to predict the added noise by minimizing:
\begin{equation}
    \mathcal{L}_{\text{DDPM}} = \mathbb{E}_{k, \boldsymbol{\tau}_0, \boldsymbol{\epsilon}} \left[ \| \boldsymbol{\epsilon} - \boldsymbol{\epsilon}_\theta(\boldsymbol{\tau}^{(k)}, k, \mathbf{c}) \|^2 \right],
\end{equation}
where $k \sim \text{Uniform}(1, K)$ and $\mathbf{c} = (\mathbf{z}_t, \mathbf{s}_{t-4:t})$ is the perception-proprioception conditioning. By training on collision-free trajectories from hindsight labeling, the diffusion policy learns to generate motions that are implicitly biased toward safe regions of the trajectory space. Combined with the motion prior from human data, this yields a planner capable of proposing diverse, feasible whole-body trajectories.

\colorbold{Multimodality and Natural Motion.} A key advantage of hindsight-augmented training is multimodal behavior. Consider two obstacle configurations $\mathcal{O}_1$ and $\mathcal{O}_2$ from different base rollouts: their intersection $\mathcal{O}_1 \cap \mathcal{O}_2$ may admit multiple collision-free trajectories, each originally labeled for one configuration but valid for both. The diffusion model thus learns to associate similar perceptual contexts with diverse trajectory modes. Furthermore, since our command space is grounded in LAFAN1~\citep{harvey2020lafan1}, the generated trajectories inherit natural, human-like characteristics. Together, this yields a planner proposing diverse, feasible whole-body trajectories for test-time steering.

\subsection{Test-Time Steering}
\label{sec:steering}

While the trained diffusion policy produces generally safe trajectories, we further refine behavior at test time using lightweight steering mechanisms. This enables adapting to new objectives without retraining.

\colorbold{Monte Carlo Sampling.} At each planning step, we generate $N$ candidate trajectories from the diffusion policy and select the best one according to a user-defined objective function $f(\mathbf{s}_t, \mathbf{z}_t, \boldsymbol{\tau})$:
\begin{equation}
    \boldsymbol{\tau}^* = \argmax_{\boldsymbol{\tau}^{(i)}, i \in \{1, \ldots, N\}} f(\mathbf{s}_t, \mathbf{z}_t, \boldsymbol{\tau}^{(i)}),
    \label{eq:mc_steering}
\end{equation}
where $\mathbf{s}_t$ is the proprioceptive state, $\mathbf{z}_t$ is the perception embedding, and $\boldsymbol{\tau}^{(i)}$ is the $i$-th sampled trajectory (action chunk). This formulation admits any objective that depends on the current state and candidate trajectory. For instance, simple objectives such as commanding the robot to turn toward a specific heading or maintaining a desired upper-body posture can be defined analytically and tuned online without retraining. Table~\ref{tab:steering_objectives} lists the analytical objectives used in our experiments.

\begin{table}[t]
    \centering
    \caption{Examples of plug-in analytical steering objectives. Each objective can replace $f$ in Eq.~\eqref{eq:mc_steering} without retraining the diffusion policy.}
    \label{tab:steering_objectives}
    \small
    \begin{adjustbox}{max width=\linewidth}
    \begin{tabular}{ll}
        \toprule
        Objective & Score $f(\boldsymbol{\tau})$ \\
        \midrule
        Base speed & $\exp(-\|\mathbf{v}^{b}_{t:t+H}-\mathbf{v}^{*}\|_2/\beta)$ \\
        Turning direction & $\exp(-\|\psi_{t:t+H}-\psi^{*}\|_2/\beta)$ \\
        Base goal reaching & $\exp(-\|\mathbf{p}^{b}_{t+H}-\mathbf{p}_{\mathrm{goal}}\|_2/\beta)$ \\
        Base height & $\exp(-\|h_{t:t+H}-h^{*}\|_2/\beta)$ \\
        Left arm lifting & $\exp(-\|\mathbf{p}^{l}_{t:t+H}-\mathbf{p}^{l*}\|_2/\beta)$ \\
        Right arm lifting & $\exp(-\|\mathbf{p}^{r}_{t:t+H}-\mathbf{p}^{r*}\|_2/\beta)$ \\
        \bottomrule
    \end{tabular}%
    \end{adjustbox}
\end{table}

\colorbold{Learned Collision-Aware Q-Function.} For more complex objectives such as collision avoidance, we learn a Q-function through offline policy evaluation on the hindsight-labeled rollouts. The high-level MDP runs at 20\,Hz with state $\mathbf{c}_t=(\mathbf{s}_t,\mathbf{z}_t)$ and action $\boldsymbol{\tau}_{t:t+H}$, the trajectory chunk proposed by the diffusion model. The environment transition is induced by executing the selected chunk through the low-level controller. We assign a chunk-level reward
\begin{equation}
    r_t = 0.1\,V_{\mathrm{swept}}(\boldsymbol{\tau}_{t:t+H}) - V_{\mathrm{collision}}(\boldsymbol{\tau}_{t:t+H}, \mathcal{O}),
\end{equation}
where $V_{\mathrm{swept}}$ rewards progress through free swept volume and $V_{\mathrm{collision}}$ penalizes intersection with obstacles. Because the dataset consists of complete rollouts, we estimate returns with offline GAE~\citep{schulman2016gae} using $\lambda=0.95$ and $\gamma=0.99$, then regress $Q_\phi(\mathbf{c}_t,\boldsymbol{\tau}_{t:t+H})$ to these targets. At test time, this learned Q-function is used only to score sampled candidates; it does not modify the diffusion prior or require task-specific retraining of the policy.

\colorbold{Comparison with Guidance Methods.} We compare Monte Carlo steering with two prominent alternatives. \textit{Classifier-based guidance}~\citep{dhariwal2021diffusion} modifies the score using classifier gradients:
\begin{equation}
    \tilde{\boldsymbol{\epsilon}}_\theta = \boldsymbol{\epsilon}_\theta(\boldsymbol{\tau}^{(k)}, k, \mathbf{c}) - s \cdot \nabla_{\boldsymbol{\tau}^{(k)}} \log p_\phi(y \mid \boldsymbol{\tau}^{(k)}).
\end{equation}
This requires differentiable objectives and careful tuning of $s$ at each noise level. \textit{Classifier-free guidance} (CFG)~\citep{ho2022classifier} co-trains conditional and unconditional models:
\begin{equation}
    \tilde{\boldsymbol{\epsilon}}_\theta = \boldsymbol{\epsilon}_\theta(\boldsymbol{\tau}^{(k)}, k, \mathbf{c}) + w \cdot \left( \boldsymbol{\epsilon}_\theta(\boldsymbol{\tau}^{(k)}, k, \mathbf{c}, y) - \boldsymbol{\epsilon}_\theta(\boldsymbol{\tau}^{(k)}, k, \mathbf{c}) \right).
\end{equation}
However, CFG requires retraining with each new objective label $y$. In contrast, our Monte Carlo steering operates purely at test time by selecting among samples from the learned distribution according to $f$, avoiding out-of-distribution artifacts from gradient-based guidance.

\section{Experiments}
\label{sec:experiments}

We evaluate \method{} in both simulation and real-world deployment on the Unitree G1 humanoid robot. Our experiments address the following questions:
\begin{enumerate}
    \item Can \method{} successfully solve the obstacle avoidance problem in both simulation and the real world?
    \item How does hindsight labeling benefit the data? Is it scalable and how well does it generalize?
    \item How does test-time steering benefit the framework? Is it easy to integrate as a plug-in module?
\end{enumerate}

\subsection{Experimental Setup}

\colorbold{Task Design.} We use Isaac Gym~\citep{makoviychuk2021isaacgym} for physics simulation with a Unitree G1 humanoid model (29 DoF). We evaluate on five obstacle avoidance tasks: \textit{Walk sideways}---navigating through a narrow lateral gap between two close obstacles; \textit{Ducking}---lowering the body to pass under a head-height horizontal bar; \textit{Lift hand to pass}---raising one arm to clear an arm-height obstacle; \textit{Fast dynamic obstacle}---avoiding a ball swinging on a rope; and \textit{Slow dynamic obstacle}---upper-body adjustments to avoid a slowly moving obstacle while the base remains stationary.

\colorbold{LiDAR Simulation.} We simulate a spinning LiDAR sensor mounted on the robot's torso, providing 360-degree coverage with 16 vertical channels~\citep{wang2025omniperception, wang2025safecomfortable}. Point clouds are generated at 10\,Hz with realistic noise characteristics.

\colorbold{Training Details.} We train the low-level controller with PPO~\citep{schulman2017ppo} for 50K iterations. The perception encoder is trained for 200K steps on $8\times$ L40S GPUs (6 hours) and produces a 32-dimensional embedding. The high-level diffusion model is trained for 500K steps on $8\times$ L40S GPUs (5 hours). Test-time steering uses $N=32$ Monte Carlo samples. Unless otherwise stated, all static obstacle tasks use the same high-level diffusion policy and perception encoder; adaptation across speed, heading, and posture preferences is performed only through test-time steering. Dynamic obstacle tasks use a separately trained high-level policy because the 10\,Hz LiDAR stream is insufficient for high-frequency dynamic obstacle state estimation.

\colorbold{Sim2Sim Validation.} To validate simulation fidelity, we perform sim2sim transfer by deploying policies trained in Isaac Gym to MuJoCo~\citep{todorov2012mujoco}, confirming consistent behavior across physics engines.

\colorbold{Baselines.} We compare the following methods:
\begin{itemize}
    \item \textbf{End2End Diffusion}~\citep{huang2024diffuseloco, huang2025diffusecloc}: Uses diffusion policy outputs directly for control without the low-level RL policy, to validate the necessity of the hierarchical design
    \item \textbf{\method{} (no steering)}: Full hierarchical framework with hindsight training but no test-time steering
    \item \textbf{\method{} (full)}: Complete framework with both hindsight training and Monte Carlo steering
\end{itemize}

\colorbold{Metrics.} We evaluate using: \textit{Success Rate}---percentage of episodes reaching the goal without falling or collision; and \textit{Collision Rate}---the obstacle set is a wall of 180 cubes, as shown in Figure~\ref{fig:sim_results}, and the collision rate is the percentage of cubes that are contacted by any part of the robot body.

\subsection{Obstacle Avoidance Performance (Q1)}

We first evaluate whether \method{} can successfully solve the obstacle avoidance problem in both simulation and real-world settings.

\subsubsection{Simulation Results}
\begin{table}[t]
    \centering
    \caption{Success rate (Succ.) and collision rate (Coll.) across different obstacle avoidance tasks (mean $\pm$ std over 3 seeds). End2End Diffusion uses diffusion outputs directly without the whole-body tracking RL policy, which fails to maintain balance in most locomotion tasks. \method{} with steering achieves the best overall performance.}
    \label{tab:main_results}
    \begin{adjustbox}{max width=\linewidth}
    \begin{NiceTabular}{lcccccc}
        \CodeBefore
        \rectanglecolor{steelgray}{1-6}{9-7}
        \Body
        \toprule
        \multirow{2}{*}{Task} & \multicolumn{2}{c}{End2End Diffusion} & \multicolumn{2}{c}{\method{} (no steer)} & \multicolumn{2}{c}{\method{} (full)} \\
        \cmidrule(lr){2-3} \cmidrule(lr){4-5} \cmidrule(lr){6-7}
        & Succ.$\uparrow$ & Coll.$\downarrow$ & Succ.$\uparrow$ & Coll.$\downarrow$ & Succ.$\uparrow$ & Coll.$\downarrow$ \\
        \midrule
        Walk sideways & 8.2\%{\tiny$\pm$2.1} & 4.1\%{\tiny$\pm$1.3} & 72.5\%{\tiny$\pm$3.8} & 12.3\%{\tiny$\pm$2.9} & \textbf{88.4}\%{\tiny$\pm$2.4} & \textbf{2.2}\%{\tiny$\pm$0.8} \\
        Ducking & 5.6\%{\tiny$\pm$1.8} & 3.2\%{\tiny$\pm$1.1} & 68.3\%{\tiny$\pm$4.2} & 16.8\%{\tiny$\pm$3.5} & \textbf{85.7}\%{\tiny$\pm$2.9} & \textbf{2.5}\%{\tiny$\pm$1.1} \\
        Lift hand to pass & 15.4\%{\tiny$\pm$4.6} & 20.8\%{\tiny$\pm$8.7} & 74.1\%{\tiny$\pm$3.5} & 10.5\%{\tiny$\pm$2.8} & \textbf{91.2}\%{\tiny$\pm$2.1} & \textbf{1.3}\%{\tiny$\pm$0.3} \\
        Fast dynamic obs. & 6.3\%{\tiny$\pm$1.9} & - & 65.8\%{\tiny$\pm$4.6} & - & \textbf{82.6}\%{\tiny$\pm$3.2} & -\\
        Slow dynamic obs. & 51.5\%{\tiny$\pm$4.3} & - & 81.2\%{\tiny$\pm$3.1} & - & \textbf{93.4}\%{\tiny$\pm$1.8} & - \\
        \midrule
        Average & 17.4\%{\tiny$\pm$2.9} & 9.4\%{\tiny$\pm$3.7} & 72.4\%{\tiny$\pm$3.8} & 13.2\%{\tiny$\pm$3.1} & \textbf{88.3}\%{\tiny$\pm$2.5} & \textbf{2.0}\%{\tiny$\pm$0.5} \\
        \bottomrule
    \end{NiceTabular}%
    \end{adjustbox}
\end{table}

Table~\ref{tab:main_results} presents our main simulation results across five obstacle avoidance tasks. The collision rate is missing for Fast dynamic obstacle and Slow dynamic obstacle tasks because there is only one obstacle in the scene, so an unsuccessful episode means the obstacle hit the robot. Key observations:

\colorbold{End-to-end diffusion control fails without the RL policy.} Using diffusion outputs directly for control---i.e., removing the low-level RL policy---fails on locomotion-intensive tasks (5--15\% success). The robot frequently falls when agility is required, e.g., turning while sidestepping or squatting to duck under obstacles. Performance is higher on \textit{Lift hand to pass}, which largely reduces to straight-line walking with upper-body motion. The main exception is \textit{Slow dynamic obstacle}: the robot can remain stationary and move only its arms to avoid slowly approaching objects, achieving 51.5\% success. This also explains why End2End Diffusion can show a low collision rate despite low success: many failures occur before the robot reaches the obstacle because the generated motion is dynamically unstable rather than overly conservative.

\colorbold{Steering significantly improves performance.} Comparing \method{} with and without steering reveals consistent improvements across all tasks. Steering reduces collision rates by 9--14\% absolute while boosting success rates by 12--18\%, demonstrating the effectiveness of Monte Carlo sampling for test-time trajectory selection. The objective function $f(\cdot)$ used for collision-aware steering is the learned Q-function described in Section~\ref{sec:steering}.

\colorbold{\method{} enables diverse whole-body avoidance.} Our full method achieves an 88.3\% average success rate, demonstrating effective whole-body coordination across diverse scenarios, even for dynamic obstacle avoidance.

\subsubsection{Real-World Deployment}
\begin{figure}[t]
    \centering
    \includegraphics[width=0.85\linewidth]{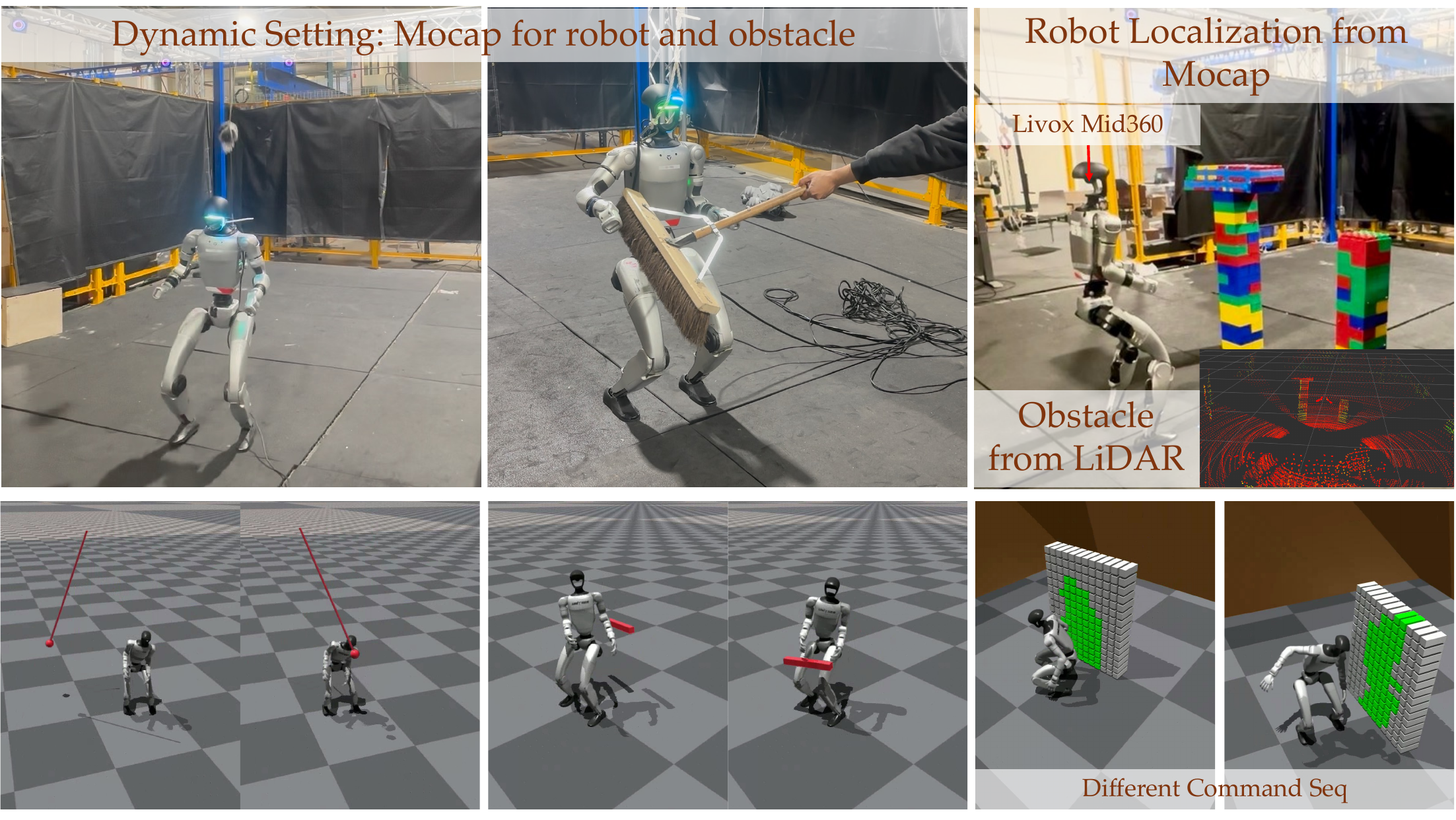}
    \caption{Simulation results and real-world settings for five obstacle avoidance tasks.}
    \label{fig:sim_results}
\end{figure}

We deploy \method{} on a physical Unitree G1 equipped with a Livox Mid-360 LiDAR sensor (Figure~\ref{fig:sim_results}). The LiDAR produces point clouds at 10\,Hz, and the perception encoder runs at the same rate. The high-level diffusion policy runs at 20\,Hz and outputs a 5-step action chunk, which is tracked by the low-level RL controller running at 50\,Hz on an onboard NVIDIA Jetson Orin. Chunks are executed in a receding-horizon manner: a new high-level plan replaces the previous one every 0.05\,s, while the low-level controller interpolates and tracks the current waypoint buffer. For static obstacle avoidance tasks, obstacle perception comes entirely from the LiDAR-based perception module. Motion capture (MoCap) is used for robot localization and for specifying dynamic obstacle states, but not for static obstacle perception.
\begin{table}[t]
    \centering
    \caption{Real-world performance on the Unitree G1 across 5 trials per scenario.}
    \label{tab:real_world}
    \begin{adjustbox}{max width=\linewidth}
    \begin{tabular}{lccccc}
        \toprule
        & Walk sideways & Ducking & Lift hand & Fast dyn. & Slow dyn. \\
        \midrule
        Succ & 4/5 & 3/5 & 4/5 & 3/5 & 4/5 \\
        \bottomrule
    \end{tabular}
    \end{adjustbox}
\end{table}
Table~\ref{tab:real_world} shows that \method{} successfully transfers to the real world. The robot exhibits the same whole-body avoidance behaviors (ducking, sidestepping) observed in simulation. Performance degrades gracefully in more challenging combined scenarios, with failure cases primarily due to LiDAR occlusion from extreme body poses and sim-to-real tracking mismatch for more agile motions. These observations suggest that perception and tracking fidelity, rather than high-level mode diversity, are the main bottlenecks in the current hardware deployment.

\subsection{Analysis of Hindsight Labeling (Q2)}

We investigate three aspects of hindsight labeling: (1) its effectiveness and scalability, (2) the role of cost augmentation ratio, and (3) generalization to unseen scenarios.

\begin{figure}[t]
    \centering
    \includegraphics[width=0.9\linewidth]{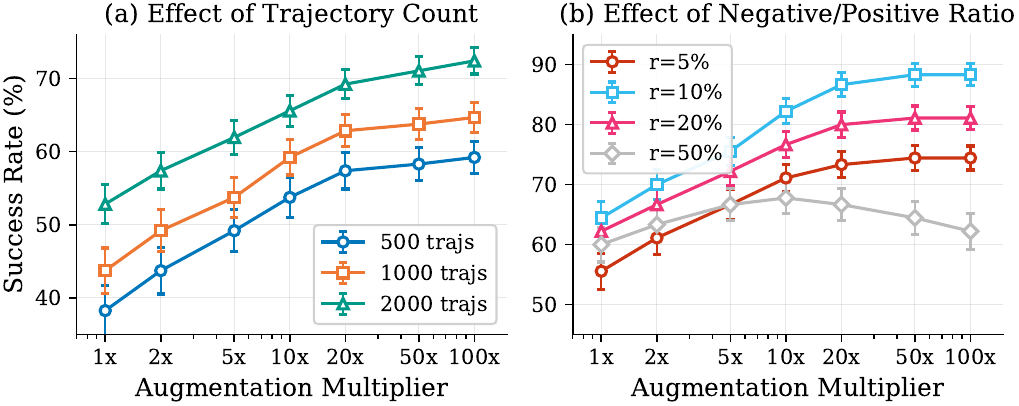}
    \caption{Ablation study on hindsight labeling. (a) Trajectory count and data augmentation, without test-time steering: more base trajectories and higher augmentation multipliers yield higher success rates. (b) 
    Positive/negative augmentation ratio, with test-time steering: a 10\% ratio achieves optimal performance; higher ratios (50\%) introduce too many negative samples, degrading performance at high augmentation levels. Error bars indicate standard deviation over 5 seeds.}
    \label{fig:ablation_hindsight}
\end{figure}

\colorbold{Hindsight labeling is effective and scalable.} Figure~\ref{fig:ablation_hindsight}(a) shows that both increasing the number of base trajectories and applying data augmentation consistently improve success rate. The augmentation process requires only simulation rollouts with automatic collision annotation, making it straightforward to scale data collection.

\colorbold{Cost augmentation ratio is critical.} Hindsight labeling augments the dataset with both safe (positive) and collision (negative) trajectory samples. The negative samples provide a correction signal that guides the Q-function away from collision-prone behaviors. However, the ratio of negative to positive samples significantly impacts performance. As shown in Figure~\ref{fig:ablation_hindsight}(b), a 10\% negative ratio achieves optimal performance by balancing safety awareness with behavioral diversity. Excessive negative samples (50\%) undermine the quality of the training distribution, leading to overly conservative policies that degrade at high augmentation levels. Notably, when combined with test-time steering, performance saturates earlier (around $50\times$ augmentation) compared to the no-steering setting. This indicates that steering effectively exploits the Q-function learned from hindsight data, reducing the data requirements for the diffusion policy.

\begin{figure}[t]
    \centering
    \includegraphics[width=0.8\linewidth]{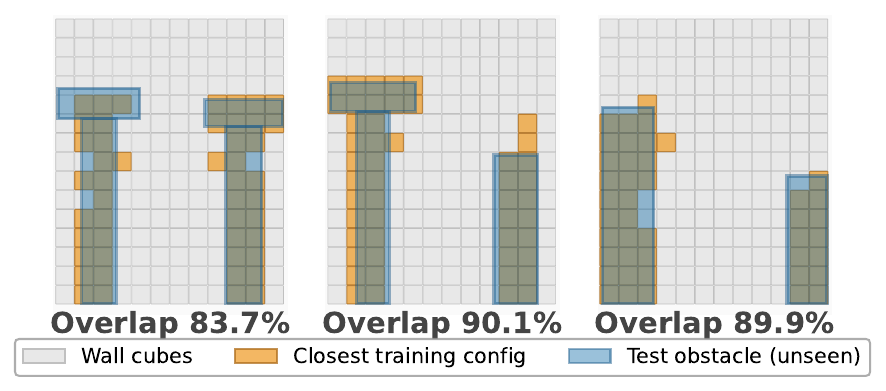}
    \caption{Training obstacles (orange) and held-out test obstacles (blue) for representative wall configurations. The closest training configuration overlaps 84--90\% with the test scene, but the exact wall positions, yaw angles, and gap sizes are never seen during training.}
    \label{fig:generalization_obstacles}
\end{figure}

\colorbold{Generalization to unseen scenarios.} The results of Figure~\ref{fig:ablation_hindsight} are evaluated on held-out obstacle configurations disjoint from those used during training. Here, ``unseen'' means new wall positions, yaw angles, gap locations, and gap sizes sampled from the same distribution family with a separate random seed. Figure~\ref{fig:generalization_obstacles} visualizes the closest training configurations for representative held-out scenes. Although the closest training obstacle overlaps 84--90\% with the test configuration, the remaining geometric differences are sufficient to change the feasible whole-body maneuver. This supports the claim that the perception encoder learns a spatial representation useful for obstacle reasoning rather than memorizing exact obstacle grids. We do not claim full out-of-distribution generalization to arbitrary furniture-like scenes; broader scene diversity remains a limitation and a natural direction for scaling the hindsight pipeline.
\begin{figure}[t]
    \centering
    \includegraphics[width=0.85\linewidth]{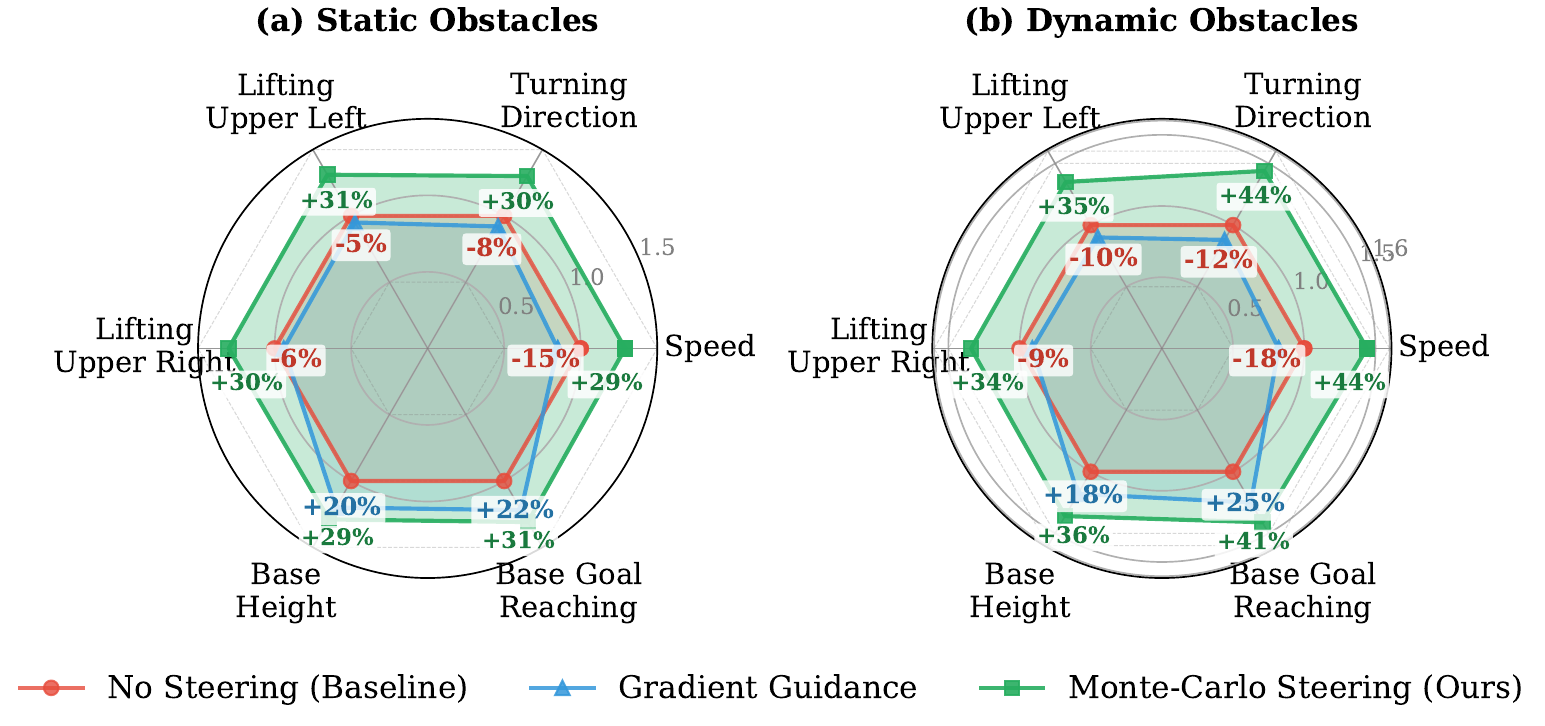}
    \caption{Steering improvement across six objectives. Radar charts compare baseline (no steering) versus different steering methods for (a) static and (b) dynamic obstacles.}
    \label{fig:steering_radar}
\end{figure}
\subsection{Analysis of Test-Time Steering (Q3)}

We analyze test-time steering along three dimensions: (1) plug-in flexibility with simple objectives, (2) effectiveness across different objectives, and (3) coupling between steering objectives.

\colorbold{Real deployment configuration.} In our real-world experiments, the diffusion policy generates a batch of action chunks with chunk size 5 and batch size $N=32$. By leveraging GPU-accelerated ONNX inference~\citep{onnxruntime}, batch size $N=32$ runs at essentially the same speed as $N=1$, enabling effective Monte Carlo steering without sacrificing the 20\,Hz policy frequency required for high-speed obstacle avoidance. Appendix~\ref{appendix:inference_latency} gives the full latency benchmark.

\colorbold{Any simple objective can be plugged in.} Our steering framework supports any objective function that can be evaluated from the candidate trajectory $\boldsymbol{\tau}$. For example, to bias the robot toward a desired posture, we use
\[
f(\boldsymbol{\tau}) = \exp\!\left(-\left\lVert q^{[j]}_{t:t+H} - q^{[j]}_{\text{des},t:t+H}\right\rVert_2 / \beta\right),
\]
where $[j]$ indexes the joints relevant to the objective and $\beta$ is a normalization hyperparameter. We evaluate six such objectives: base speed, turning direction, base goal reaching, base height, and left/right arm lifting. This plug-and-play design requires no retraining: users specify the target posture, and Monte Carlo sampling selects the trajectory that best satisfies it. Compared to gradient-based guidance, which requires careful tuning per objective, sampling-based selection is robust across objectives and less sensitive to hyperparameter choices.

\colorbold{Objective dimensionality affects steering effectiveness.} Figure~\ref{fig:steering_radar} shows that, with dynamic obstacles, base-related objectives (speed, turning direction, goal reaching) achieve larger improvements (41--44\%) than upper-body objectives (arm lifting, 34--35\%). With static obstacles, all six objectives improve by a similar 29--31\%. Lower-dimensional objectives (1--2 dimensions for base control) are easier to optimize via sampling, whereas higher-dimensional objectives (arm configurations) are more sensitive for gradient-based guidance. The base objectives gain more with dynamic obstacles because the feasible non-collision set is larger and the diffusion policy's output distribution is wider, providing more diverse candidates for selection.

\begin{figure}[t]
    \centering
    \includegraphics[width=0.7\linewidth]{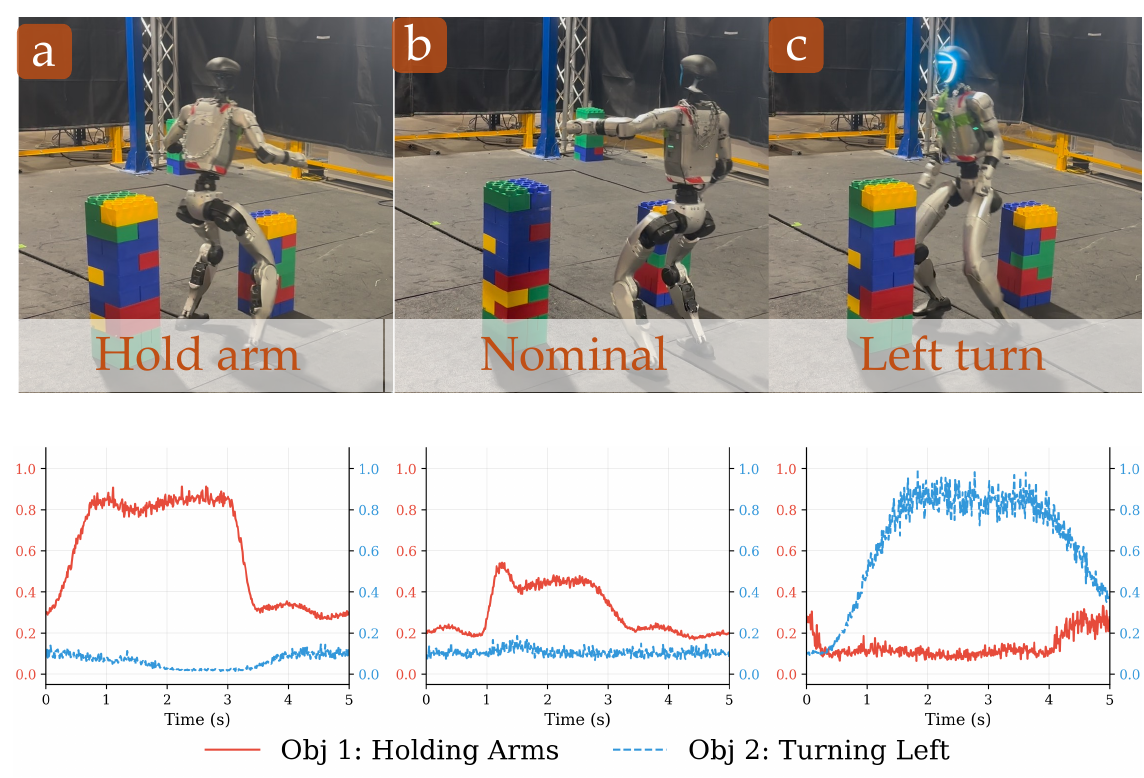}
    \caption{Comparison of three steering configurations: (a) holding both arms, (b) nominal without steering, and (c) turning left. Each panel shows the values of both objectives over time.}
    \label{fig:steering_online}
\end{figure}

\colorbold{Steering induces distinct avoidance modes.} Figure~\ref{fig:steering_online} compares three steering configurations: holding both arms (a), nominal without steering (b), and turning left (c). Two key observations emerge. First, steering objectives are not orthogonal---when steering to hold both arms (a), the turning objective, which is not steered, drops below its nominal value, indicating that the diffusion policy couples these behaviors rather than treating them as independent joint-level controls. Second, steering does not merely override specific joints but guides the robot into fundamentally different avoidance strategies. The hold-arm objective induces lateral sidestepping behavior, while the left-turning objective triggers upper-body adjustments with accompanying postural changes. 

This holistic response demonstrates that Monte Carlo steering operates at the level of motion modes rather than individual degrees of freedom.

\section{Conclusion}
\label{sec:conclusion}

We presented \method{}, a hierarchical diffusion-based framework for whole-body obstacle avoidance in humanoid robots. By combining a perception-conditioned diffusion planner with a low-level RL locomotion controller, our approach enables natural, multimodal avoidance behaviors that exploit the full kinematic capabilities of a humanoid body. Hindsight data collection automatically generates collision-aware supervision without manual labeling, while test-time Monte Carlo steering enables robust navigation with minimal reward engineering.

Experiments on the Unitree G1 demonstrate significant improvements over baselines, with diverse avoidance strategies---sidestepping, ducking, twisting---emerging naturally from the learned motion prior.

\colorbold{Limitations.} Our framework has two primary limitations. First, test-time steering currently operates with a limited batch size ($N=32$ Monte Carlo samples) due to running the high-level diffusion policy at 20\,Hz; larger batch sizes could improve steering capability for more challenging environments. Second, our perception module generalizes to held-out obstacle configurations within our training distribution, but may struggle with highly diverse unseen environments that differ significantly from training scenarios.

\colorbold{Future Work.} Promising directions include deeper integration with foundation navigation models, where hindsight labeling combined with procedural scene generation could produce more diverse training environments. Additional directions include extending \method{} to loco-manipulation tasks, incorporating vision or language for semantic navigation, and investigating efficient diffusion architectures to increase sampling throughput.



\bibliographystyle{plainnat}
\bibliography{refs}

@inproceedings{gu2024rt_trajectory,
  title={\href{https://arxiv.org/abs/2311.01977}{RT-Trajectory: Robotic Task Generalization via Hindsight Trajectory Sketches}},
  author={Gu, Jiayuan and Kirmani, Sean and Wohlhart, Paul and Lu, Yao and Arenas, Montserrat Gonzalez and Rao, Kanishka and Yu, Wenhao and Fu, Chuyuan and Gopalakrishnan, Keerthana and Xu, Zhuo and Sundaresan, Priya and Xu, Peng and Su, Hao and Hausman, Karol and Finn, Chelsea and Vuong, Quan and Xiao, Ted},
  booktitle={International Conference on Learning Representations (ICLR)},
  year={2024}
}

@inproceedings{ze20243ddp,
  title={\href{https://arxiv.org/abs/2403.03954}{3D Diffusion Policy: Generalizable Visuomotor Policy Learning via Simple 3D Representations}},
  author={Ze, Yanjie and Zhang, Gu and Zhang, Kangning and Hu, Chenyuan and Wang, Muhan and Xu, Huazhe},
  booktitle={Robotics: Science and Systems (RSS)},
  year={2024}
}

@article{radosavovic2024realworld,
  title={\href{https://arxiv.org/abs/2303.03381}{Real-World Humanoid Locomotion with Reinforcement Learning}},
  author={Radosavovic, Ilija and Xiao, Tete and Zhang, Bike and Darrell, Trevor and Malik, Jitendra and Sreenath, Koushil},
  journal={Science Robotics},
  volume={9},
  year={2024}
}

@article{zhang2025falcon,
  title={\href{https://arxiv.org/abs/2505.06776}{FALCON: Learning Force-Adaptive Humanoid Loco-Manipulation}},
  author={Zhang, Yuanhang and Yuan, Yifu and Gurunath, Prajwal and Gupta, Ishita and Omidshafiei, Shayegan and Agha-mohammadi, Ali-akbar and Vazquez-Chanlatte, Marcell and Pedersen, Liam and He, Tairan and Shi, Guanya},
  journal={arXiv preprint arXiv:2505.06776},
  year={2025}
}

@article{xue2026humanoidpf,
  title={\href{https://arxiv.org/abs/2601.16035}{Collision-Free Humanoid Traversal in Cluttered Indoor Scenes}},
  author={Xue, Han and Liang, Sikai and Zhang, Zhikai and Zeng, Zicheng and Liu, Yun and Lian, Yunrui and Wang, Jilong and Liu, Qingtao and Shi, Xuesong and Yi, Li},
  journal={IEEE Robotics and Automation Letters},
  volume={11},
  year={2026}
}

@article{liao2025beyondmimic,
  title={\href{https://arxiv.org/abs/2508.08241}{BeyondMimic: From Motion Tracking to Versatile Humanoid Control via Guided Diffusion}},
  author={Liao, Qiayuan and Truong, Takara E. and Huang, Xiaoyu and Gao, Yuman and Tevet, Guy and Sreenath, Koushil and Liu, C. Karen},
  journal={arXiv preprint arXiv:2508.08241},
  year={2025}
}

@inproceedings{cai2025navdp,
  title={\href{https://arxiv.org/abs/2505.08712}{NavDP: Learning Sim-to-Real Navigation Diffusion Policy with Privileged Information Guidance}},
  author={Cai, Wenzhe and Peng, Jiaqi and Yang, Yuqiang and Zhang, Yujian and Wei, Meng and Wang, Hanqing and Chen, Yilun and Wang, Tai and Pang, Jiangmiao},
  booktitle={IEEE International Conference on Robotics and Automation (ICRA)},
  year={2026}
}

@article{lin2025hwcloco,
  title={\href{https://arxiv.org/abs/2503.00923}{HWC-Loco: A Hierarchical Whole-Body Control Approach to Robust Humanoid Locomotion}},
  author={Lin, Sixu and Qiao, Guanren and Tai, Yunxin and Li, Ang and Jia, Kui and Liu, Guiliang},
  journal={arXiv preprint arXiv:2503.00923},
  year={2025}
}

@inproceedings{zhang2025naturalgmp,
  title={\href{https://arxiv.org/abs/2503.09015}{Natural Humanoid Robot Locomotion with Generative Motion Prior}},
  author={Zhang, Haodong and Zhang, Liang and Chen, Zhenghan and Chen, Lu and Wang, Yue and Xiong, Rong},
  booktitle={IEEE/RSJ International Conference on Intelligent Robots and Systems (IROS)},
  year={2025}
}

@inproceedings{kalaria2025dreamcontrol,
  title={\href{https://arxiv.org/abs/2509.14353}{DreamControl: Human-Inspired Whole-Body Humanoid Control for Scene Interaction via Guided Diffusion}},
  author={Kalaria, Dvij and Harithas, Sudarshan S and Katara, Pushkal and Kwak, Sangkyung and Bhagat, Sarthak and Sastry, Shankar and Sridhar, Srinath and Vemprala, Sai and Kapoor, Ashish and Huang, Jonathan Chung-Kuan},
  booktitle={IEEE International Conference on Robotics and Automation (ICRA)},
  year={2026}
}

@article{wang2025safecomfortable,
  title={\href{https://arxiv.org/abs/2508.07611}{End-to-End Humanoid Robot Safe and Comfortable Locomotion Policy}},
  author={Wang, Zifan and Yang, Xun and Zhao, Jianzhuang and Zhou, Jiaming and Ma, Teli and Gao, Ziyao and Ajoudani, Arash and Liang, Junwei},
  journal={arXiv preprint arXiv:2508.07611},
  year={2025}
}

@article{wang2025omniperception,
  title={\href{https://arxiv.org/abs/2505.19214}{Omni-Perception: Omnidirectional Collision Avoidance for Legged Locomotion in Dynamic Environments}},
  author={Wang, Zifan and Ma, Teli and Jia, Yufei and Yang, Xun and Zhou, Jiaming and Ouyang, Wenlong and Zhang, Qiang and Liang, Junwei},
  journal={arXiv preprint arXiv:2505.19214},
  year={2025},
}

@inproceedings{huang2024diffuseloco,
  title={\href{https://arxiv.org/abs/2404.19264}{DiffuseLoco: Real-Time Legged Locomotion Control with Diffusion from Offline Datasets}},
  author={Huang, Xiaoyu and Chi, Yufeng and Wang, Ruofeng and Li, Zhongyu and Peng, Xue Bin and Shao, Sophia and Nikolic, Borivoje and Sreenath, Koushil},
  booktitle={Conference on Robot Learning (CoRL)},
  year={2024}
}

@article{huang2025diffusecloc,
  title={\href{https://arxiv.org/abs/2503.11801}{Diffuse-CLoC: Guided Diffusion for Physics-based Character Look-ahead Control}},
  author={Huang, Xiaoyu and Truong, Takara and Zhang, Yunbo and Yu, Fangzhou and Sleiman, Jean Pierre and Hodgins, Jessica and Sreenath, Koushil and Farshidian, Farbod},
  journal={ACM Transactions on Graphics},
  volume={44},
  year={2025}
}

@inproceedings{ze2024humanoidmanipulation,
  title={\href{https://arxiv.org/abs/2410.10803}{Generalizable Humanoid Manipulation with 3D Diffusion Policies}},
  author={Ze, Yanjie and Chen, Zixuan and Wang, Wenhao and Chen, Tianyi and He, Xialin and Yuan, Ying and Peng, Xue Bin and Wu, Jiajun},
  booktitle={IEEE/RSJ International Conference on Intelligent Robots and Systems (IROS)},
  year={2025}
}

@inproceedings{fu2024humanplus,
  title={\href{https://arxiv.org/abs/2406.10454}{HumanPlus: Humanoid Shadowing and Imitation from Humans}},
  author={Fu, Zipeng and Zhao, Qingqing and Wu, Qi and Wetzstein, Gordon and Finn, Chelsea},
  booktitle={Conference on Robot Learning (CoRL)},
  year={2024}
}

@article{kim2024armor,
  title={\href{https://arxiv.org/abs/2412.00396}{ARMOR: Egocentric Perception for Humanoid Robot Collision Avoidance and Motion Planning}},
  author={Kim, Daehwa and Srouji, Mario and Chen, Chen and Zhang, Jian},
  journal={arXiv preprint arXiv:2412.00396},
  year={2024},
}

@article{harvey2020lafan1,
  title={\href{https://dl.acm.org/doi/10.1145/3386569.3392480}{Robust Motion In-betweening}},
  author={Harvey, F{\'e}lix G. and Yurick, Mike and Nowrouzezahrai, Derek and Pal, Christopher},
  journal={ACM Transactions on Graphics},
  volume={39},
  number={4},
  year={2020}
}

@inproceedings{peebles2023dit,
  title={\href{https://arxiv.org/abs/2212.09748}{Scalable Diffusion Models with Transformers}},
  author={Peebles, William and Xie, Saining},
  booktitle={IEEE/CVF International Conference on Computer Vision (ICCV)},
  year={2023}
}

@inproceedings{chi2023diffusionpolicy,
  title={\href{https://arxiv.org/abs/2303.04137}{Diffusion Policy: Visuomotor Policy Learning via Action Diffusion}},
  author={Chi, Cheng and Xu, Zhenjia and Feng, Siyuan and Cousineau, Eric and Du, Yilun and Burchfiel, Benjamin and Tedrake, Russ and Song, Shuran},
  booktitle={Robotics: Science and Systems (RSS)},
  year={2023}
}

@article{schulman2017ppo,
  title={\href{https://arxiv.org/abs/1707.06347}{Proximal Policy Optimization Algorithms}},
  author={Schulman, John and Wolski, Filip and Dhariwal, Prafulla and Radford, Alec and Klimov, Oleg},
  journal={arXiv preprint arXiv:1707.06347},
  year={2017}
}

@inproceedings{schulman2016gae,
  title={\href{https://arxiv.org/abs/1506.02438}{High-Dimensional Continuous Control Using Generalized Advantage Estimation}},
  author={Schulman, John and Moritz, Philipp and Levine, Sergey and Jordan, Michael and Abbeel, Pieter},
  booktitle={International Conference on Learning Representations (ICLR)},
  year={2016}
}

@inproceedings{makoviychuk2021isaacgym,
  title={\href{https://arxiv.org/abs/2108.10470}{Isaac Gym: High Performance GPU-Based Physics Simulation For Robot Learning}},
  author={Makoviychuk, Viktor and Wawrzyniak, Lukasz and Guo, Yunrong and Lu, Michelle and Storey, Kier and Macklin, Miles and Hoeller, David and Rudin, Nikita and Allshire, Arthur and Handa, Ankur and State, Gavriel},
  booktitle={NeurIPS Datasets and Benchmarks Track},
  year={2021}
}

@inproceedings{todorov2012mujoco,
  title={{MuJoCo: A Physics Engine for Model-Based Control}},
  author={Todorov, Emanuel and Erez, Tom and Tassa, Yuval},
  booktitle={IEEE/RSJ International Conference on Intelligent Robots and Systems (IROS)},
  pages={5026--5033},
  year={2012},
  organization={IEEE}
}

@inproceedings{ames2019cbf,
  title={\href{https://arxiv.org/abs/1903.11199}{Control Barrier Functions: Theory and Applications}},
  author={Ames, Aaron D. and Coogan, Samuel and Egerstedt, Magnus and Notomista, Gennaro and Sreenath, Koushil and Tabuada, Paulo},
  booktitle={European Control Conference (ECC)},
  year={2019}
}

@inproceedings{andrychowicz2017her,
  title={\href{https://arxiv.org/abs/1707.01495}{Hindsight Experience Replay}},
  author={Andrychowicz, Marcin and Wolski, Filip and Ray, Alex and Schneider, Jonas and Fong, Rachel and Welinder, Peter and McGrew, Bob and Tobin, Josh and Abbeel, Pieter and Zaremba, Wojciech},
  booktitle={Advances in Neural Information Processing Systems (NeurIPS)},
  year={2017}
}

@inproceedings{dhariwal2021diffusion,
  title={\href{https://arxiv.org/abs/2105.05233}{Diffusion Models Beat {GANs} on Image Synthesis}},
  author={Dhariwal, Prafulla and Nichol, Alexander},
  booktitle={Advances in Neural Information Processing Systems},
  year={2021}
}

@article{ho2022classifier,
  title={\href{https://arxiv.org/abs/2207.12598}{Classifier-Free Diffusion Guidance}},
  author={Ho, Jonathan and Salimans, Tim},
  journal={arXiv preprint arXiv:2207.12598},
  year={2022}
}

@article{spark2025,
  title={\href{https://arxiv.org/abs/2502.03132}{SPARK: Safe Protective and Assistive Robot Kit}},
  author={Sun, Yifan and Chen, Rui and Yun, Kai S. and Fang, Yikuan and Jung, Sebin and Li, Feihan and Li, Bowei and Zhao, Weiye and Liu, Changliu},
  journal={arXiv preprint arXiv:2502.03132},
  year={2025},
  note={Presented at the IFAC Symposium on Robotics}
}

@inproceedings{he2024abs,
  title={\href{https://arxiv.org/abs/2401.17583}{Agile But Safe: Learning Collision-Free High-Speed Legged Locomotion}},
  author={He, Tairan and Zhang, Chong and Xiao, Wenli and He, Guanqi and Liu, Changliu and Shi, Guanya},
  booktitle={Robotics: Science and Systems (RSS)},
  year={2024}
}

@inproceedings{li2024lano3dp,
  title={\href{https://arxiv.org/abs/2407.00451}{Language-Guided Object-Centric Diffusion Policy for Generalizable and Collision-Aware Robotic Manipulation}},
  author={Li, Hang and Feng, Qian and Zheng, Zhi and Feng, Jianxiang and Chen, Zhaopeng and Knoll, Alois},
  booktitle={IEEE International Conference on Robotics and Automation (ICRA)},
  year={2025}
}

@article{chomp2013,
  title={{CHOMP: Covariant Hamiltonian Optimization for Motion Planning}},
  author={Zucker, Matt and Ratliff, Nathan and Dragan, Anca D. and Pivtoraiko, Mihail and Klingensmith, Matthew and Dellin, Christopher M. and Bagnell, J. Andrew and Srinivasa, Siddhartha S.},
  journal={The International Journal of Robotics Research},
  volume={32},
  number={9--10},
  year={2013}
}

@inproceedings{rail2024,
  title={\href{https://arxiv.org/abs/2409.19190}{RAIL: Reachability-Aided Imitation Learning for Safe Policy Execution}},
  author={Jung, Wonsuhk and Anthony, Dennis and Mishra, Utkarsh A. and Arachchige, Nadun Ranawaka and Bronars, Matthew and Xu, Danfei and Kousik, Shreyas},
  booktitle={IEEE International Conference on Robotics and Automation (ICRA)},
  year={2025}
}

@misc{onnxruntime,
  title={{ONNX Runtime}},
  author={{ONNX Runtime developers}},
  year={2021},
  howpublished={\url{https://onnxruntime.ai/}}
}

@inproceedings{cheng2024expressive,
  title={\href{https://arxiv.org/abs/2402.16796}{Expressive Whole-Body Control for Humanoid Robots}},
  author={Cheng, Xuxin and Ji, Yandong and Chen, Junming and Yang, Ruihan and Yang, Ge and Wang, Xiaolong},
  booktitle={Robotics: Science and Systems (RSS)},
  year={2024}
}

@inproceedings{zhuang2024parkour,
  title={\href{https://arxiv.org/abs/2406.10759}{Humanoid Parkour Learning}},
  author={Zhuang, Ziwen and Yao, Shenzhe and Zhao, Hang},
  booktitle={Conference on Robot Learning (CoRL)},
  year={2024}
}

\newpage
\appendix
\section{Notation Reference}

\label{appendix:notation}

Table~\ref{tab:notation} summarizes the key notation used throughout this paper. We follow these typographic conventions:
\begin{itemize}
    \item $\mathbf{x}$ (bold roman): vectors and matrices representing concrete numerical values
    \item $\boldsymbol{\tau}$ (bold Greek): trajectory-related quantities
    \item $\mathcal{X}$ (calligraphic): sets and spaces
\end{itemize}

\begin{table}[htbp]
\centering
\caption{Summary of notation.}
\label{tab:notation}
\small
\begin{adjustbox}{max width=\linewidth}
\begin{tabular}{cl}
\toprule
\textbf{Symbol} & \textbf{Description} \\
\midrule
\multicolumn{2}{l}{\textit{State \& Observation (vectors: $\mathbf{x}$)}} \\
$\mathbf{o}_t^{\text{obs}}$ & Exteroceptive observation (LiDAR point clouds) \\
$\mathbf{s}_t$ & Proprioceptive state \\
$\mathbf{q}_t, \dot{\mathbf{q}}_t$ & Joint positions and velocities \\
$\mathbf{p}_t^{b}, \mathbf{v}_t^{b}, h_t$ & Base position, velocity, height \\
$\mathbf{p}_t^{u}$ & Upper-body position \\
$\mathbf{a}_t$ & Motor commands (joint position targets) \\
$\mathbf{o}_t^{L}$ & Low-level policy observation (waypoint, proprioception, previous action) \\
\midrule
\multicolumn{2}{l}{\textit{Perception Encoder}} \\
$\mathbf{z}_t$ & Perception embedding: $\text{Enc}(\mathbf{o}_t^{\text{obs}}, \mathbf{z}_{t-1})$ \\
\midrule
\multicolumn{2}{l}{\textit{Trajectory \& Planning (bold Greek: $\boldsymbol{\tau}$)}} \\
$\boldsymbol{\tau}_{t:t+H}$ & Trajectory command over horizon $H$ \\
$\boldsymbol{\tau}_t$ & Single waypoint command at time $t$ \\
$\boldsymbol{\tau}^{(k)}$ & Noisy trajectory at diffusion step $k$ \\
$\boldsymbol{\tau}^*$ & Selected trajectory (highest score under $f$) \\
$\boldsymbol{\epsilon}_\theta$ & Denoising network \\
\midrule
\multicolumn{2}{l}{\textit{Policy \& Conditioning}} \\
$\pi_H$ & High-level diffusion policy \\
$\pi_L$ & Low-level RL controller \\
$\mathbf{c}$ & Diffusion conditioning: $(\mathbf{z}_t, \mathbf{s}_{t-4:t})$ \\
$H$ & Planning horizon \\
$K$ & Total diffusion steps \\
$N$ & Number of Monte Carlo samples \\
\midrule
\multicolumn{2}{l}{\textit{Reward \& Steering Objective}} \\
$r_{\text{track}}, r_{\text{locomotion}}$ & Low-level tracking and locomotion rewards \\
$f(\mathbf{s}_t, \mathbf{z}_t, \boldsymbol{\tau})$ & Steering objective (score of a candidate trajectory) \\
$V_{\mathrm{swept}}, V_{\mathrm{collision}}$ & Swept-volume progress reward and collision penalty \\
$Q_\phi(\mathbf{c}_t, \boldsymbol{\tau}_{t:t+H})$ & Learned collision-aware Q-function for steering \\
\midrule
\multicolumn{2}{l}{\textit{Sets \& Spaces (calligraphic: $\mathcal{X}$)}} \\
$\mathcal{R}(\mathbf{q}_t)$ & Robot body occupancy at configuration $\mathbf{q}_t$ \\
$\mathcal{S}$ & Swept volume: $\bigcup_{t} \mathcal{R}(\mathbf{q}_t)$ \\
$\mathcal{W}$ & Workspace region \\
$\mathcal{O}_{\text{hind}}$ & Hindsight obstacle label \\
\bottomrule
\end{tabular}
\end{adjustbox}
\end{table}

\section{Inference Latency and Deployment Pipeline}
\label{appendix:inference_latency}

Low-latency inference is critical for dynamic obstacle avoidance: the high-level policy must react quickly to moving obstacles, and Monte Carlo steering requires generating multiple trajectory candidates per planning step.
To meet these demands, we export the diffusion model to ONNX Runtime~\citep{onnxruntime} and benchmark different execution configurations on the deployment GPU.
\begin{figure}[t]
    \centering
    \begin{minipage}[t]{0.48\linewidth}
        \centering
        \includegraphics[width=0.72\linewidth]{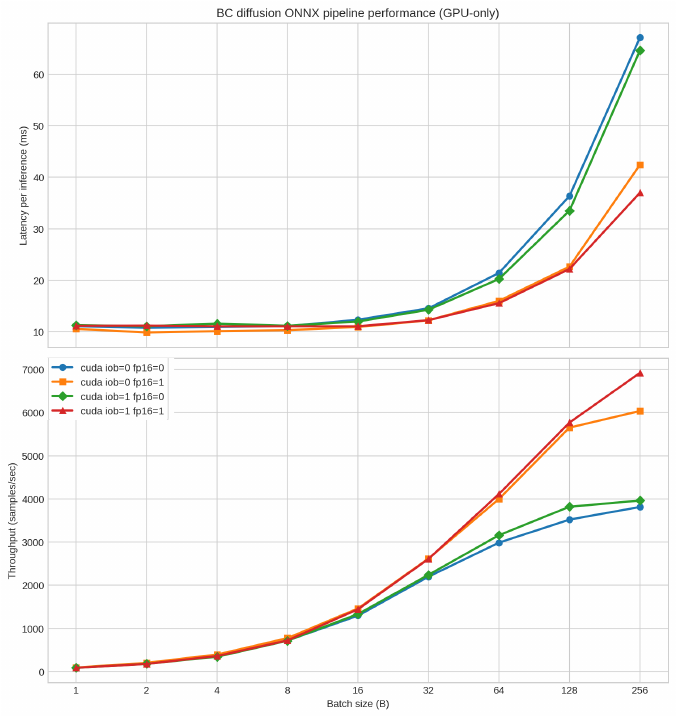}
        \caption{ONNX inference benchmark across four CUDA execution configurations (with/without IO binding, with/without FP16). \textbf{Top}: latency per inference call vs.\ batch size. \textbf{Bottom}: throughput vs.\ batch size. IO binding combined with FP16 achieves the lowest latency and highest throughput. Notably, latency remains nearly flat up to batch size ${\sim}32$ due to GPU parallelism.}
        \label{fig:gpu_performance}
    \end{minipage}\hfill
    \begin{minipage}[t]{0.48\linewidth}
        \centering
        \includegraphics[width=\linewidth]{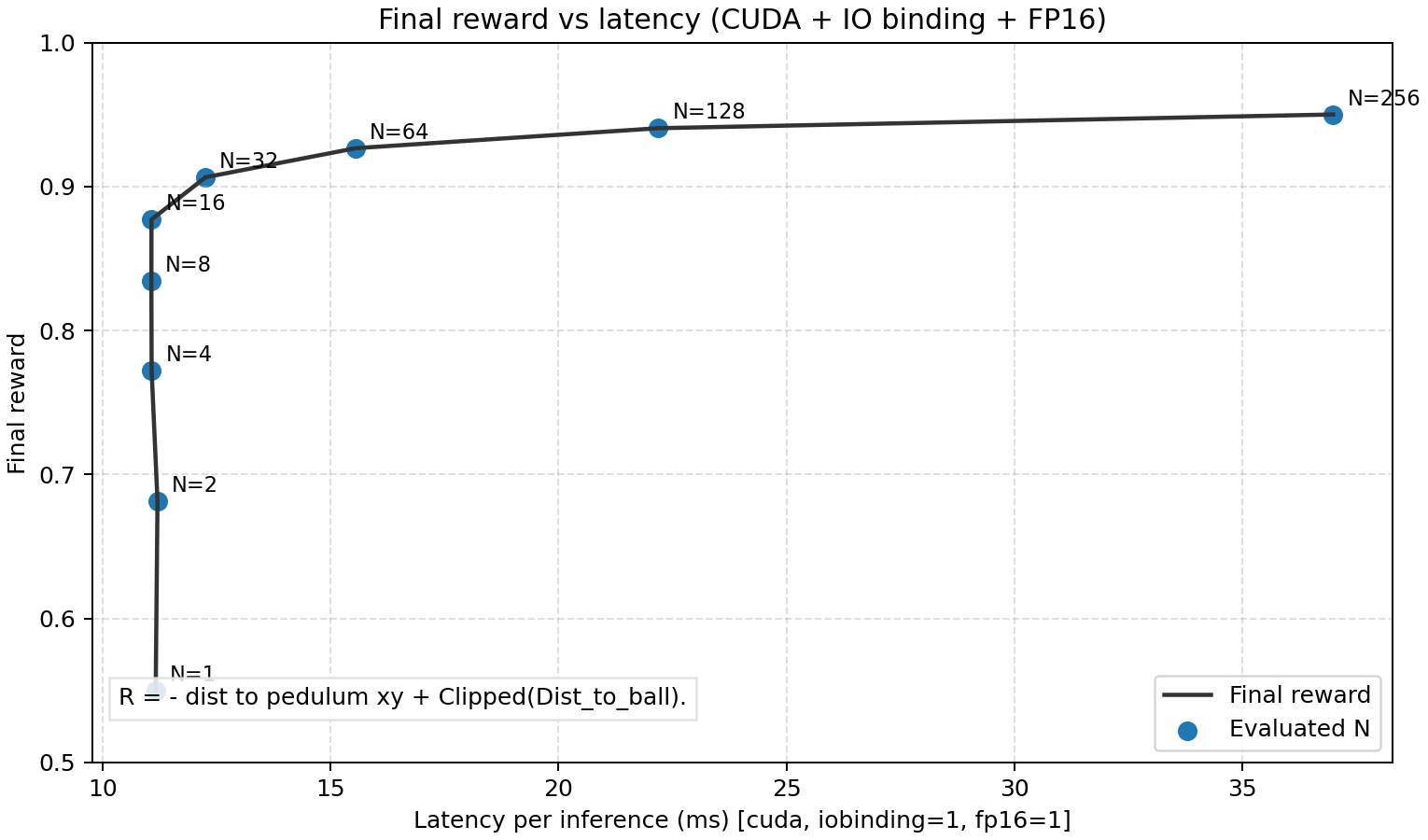}
        \caption{Final reward vs.\ inference latency under the CUDA + IO binding + FP16 configuration, annotated with batch sizes $N = 1, 2, 4, \ldots, 256$. Reward increases sharply from $N{=}1$ to $N{=}32$, then plateaus, while latency grows substantially beyond $N{=}32$. The batch size $N{=}32$ at ${\sim}13$\,ms offers the best trade-off.}
        \label{fig:reward_batch}
    \end{minipage}
\end{figure}
\Cref{fig:gpu_performance} compares four CUDA execution variants: default CUDA, CUDA with IO binding, CUDA with FP16, and CUDA with both IO binding and FP16.
IO binding eliminates redundant host-device memory copies, while FP16 halves the arithmetic precision, together yielding the lowest per-call latency and highest throughput.
A key observation is that latency stays nearly flat from batch size $N{=}1$ up to $N{\approx}32$, because the GPU's parallel execution units are not yet saturated at these batch sizes.

\Cref{fig:reward_batch} shows the reward--latency trade-off evaluated on a dynamic obstacle avoidance task.
The final reward increases sharply as the number of trajectory candidates grows from $N{=}1$ to $N{=}32$---more candidates give Monte Carlo steering a better chance of finding a high-quality trajectory---but plateaus beyond $N{=}32$ as the marginal benefit of additional samples diminishes.
Meanwhile, latency begins to increase substantially for $N > 32$.
The batch size $N{=}32$ at approximately $13$\,ms strikes the best balance between reward and latency.

Based on these results, our deployment configuration uses an action chunk length of $5$, a sample batch size of $N{=}32$, and the CUDA + IO binding + FP16 execution provider.
The high-level diffusion policy runs as fast as possible at approximately $20$\,Hz: at each planning step it generates $32$ candidate trajectory chunks, selects the best one via Monte Carlo steering, and writes it to a shared action buffer.
The low-level RL controller reads from this buffer at $50$\,Hz, converting the selected trajectory waypoint into motor commands.
This asynchronous two-rate architecture ensures that the high-level planner can exploit the full GPU throughput for trajectory sampling while the low-level controller maintains smooth, high-frequency joint tracking.

\end{document}